\documentclass[preprint,10pt,authoryear]{elsarticle}
\usepackage[margin=1in]{geometry}
\usepackage{hyperref}
\usepackage[utf8]{inputenc} % allow utf-8 input
\usepackage[T1]{fontenc}    % use 8-bit T1 fonts
\usepackage{url}            % simple URL typesetting
\usepackage{booktabs}       % professional-quality tables
\usepackage{amsfonts}       % blackboard math symbols
\usepackage{nicefrac}       % compact symbols for 1/2, etc.
\usepackage{microtype}      % microtypography
\usepackage{lipsum}
\usepackage{amsmath}
\usepackage{amssymb}
\usepackage{amsthm}
\usepackage{bm}
\usepackage{url}
\usepackage{stackengine}
\usepackage{caption}
\usepackage{makecell}
\usepackage{multirow}
\usepackage{mwe}
\usepackage{subcaption}
\usepackage{graphicx}
\usepackage{diagbox}
\usepackage{soul,color}
\usepackage[user,titleref]{zref}
\usepackage[ruled,vlined]{algorithm2e}
\usepackage{lineno}
\makeatletter
\def\ps@pprintTitle{%
  \let\@oddhead\@empty
  \let\@evenhead\@empty
  \let\@oddfoot\@empty
  \let\@evenfoot\@oddfoot
}
\makeatother

\begin{document}
% \begin{linenumbers}
\begin{frontmatter}

% \hspace{-1.5em}Title: \\
% \textbf{End-to-End Heterogeneous Graph Neural Networks for Traffic Assignment}\\

% \hfill\break \\
% \hspace{-1.5em}Authors:\\
% \hspace{-1.5em}\textbf{Tong Liu}\\
%   Department of Civil and Environmental Engineering\\
%   University of Illinois, Urbana-Champaign\\
%   tongl5@illinois.edu\\
%   https://orcid.org/0000-0002-3667-917X\\
%   \hfill\break \\% this is a way to add line numbering on empty line
%   \textbf{Hadi Meidani*}\\
%   Department of Civil and Environmental Engineering\\
%   University of Illinois, Urbana-Champaign\\
%   meidani@illinois.edu\\
%   https://orcid.org/0000-0003-4651-2696 \\

% \hfill\break \\
% \hspace{-1.5em}Corresponding Author: \\
% \textbf{Hadi Meidani, meidani@illinois.edu}

\title{End-to-End Heterogeneous Graph Neural Networks for Traffic Assignment}

\author[inst1]{Tong Liu}
\author[inst1]{Hadi Meidani}
\affiliation[inst1]{organization={University of Illinois, Urbana-Champaign, Department of Civil and Environmental Engineering},%Department and Organization
            addressline={205 N Mathews Ave}, 
            city={Urbana},
            postcode={61801}, 
            state={IL},
            country={USA}}

\begin{abstract}
The traffic assignment problem is one of the significant components of traffic flow analysis for which various solution approaches have been proposed. However, deploying these approaches for large-scale networks poses significant challenges. In this paper, we leverage the power of heterogeneous graph neural networks to propose a novel end-to-end surrogate model for traffic assignment, specifically user equilibrium traffic assignment problems. Our model integrates an adaptive graph attention mechanism with auxiliary "virtual" links connecting origin-destination node pairs, This integration enables the model to capture spatial traffic patterns across different links, By incorporating the node-based flow conservation law into the overall loss function, the model ensures the prediction results in compliance with flow conservation principles, resulting in highly accurate predictions for both link flow and flow-capacity ratios. We present numerical experiments on urban transportation networks and show that the proposed heterogeneous graph neural network model outperforms other conventional neural network models in terms of convergence rate and prediction accuracy. Notably, by introducing two different training strategies, the proposed heterogeneous graph neural network model can also be generalized to different network topologies. This approach offers a promising solution for complex traffic flow analysis and prediction, enhancing our understanding and management of a wide range of transportation systems.
\end{abstract}


\end{frontmatter}

% \newpage

\section{Introduction}
\label{sec:introduction}

Traffic assignment plays a significant role in transportation network analysis. Solving the traffic assignment problem (TAP) enables a deeper understanding of traffic flow patterns and provides insights into traffic congestion and environmental impacts \citep{nie2004models,cheng2024network}. The primary objective of the TAP is to determine the traffic flow distribution and identify traffic bottlenecks on a given road network. Such analysis enables city planners to make a strategic plan for mitigating traffic congestion. For solving traffic assignment problems, there are two major formulations with different assumptions \citep{seliverstov2017development}: (1) the user equilibrium (UE) assignment and (2) the system optimum (SO) assignment. The UE assignment identifies an equilibrium state in which drivers between each origin-destination (OD) pair cannot reduce travel costs by unilaterally shifting to another route. On the other hand, in the SO assignment, drivers choose the routes collaboratively so that the total system travel time is minimized.

In addition to network analysis, traffic assignment has broader implications in various research and practical areas such as regional resilience and urban planning. For instance, city planners leverage traffic assignment to evaluate the network reliability under extreme event scenarios, including hurricanes \citep{zou2020resilience} and earthquake \citep{liu2023optimizing}. These evaluations ensure the transportation infrastructure aligns with the city's evolving needs and requirements. Furthermore, the TAPs is employed in road network design and retrofit investment optimization as a lower-level constraint in a bi-level optimization problem \citep{madadi2024hybrid}.

A preliminary and critical step in setting up TAPs involves obtaining accurate regional OD demands. While there have been significant methodological advancements in OD demand estimation \citep{tang2021dynamic}, the existing practical challenges lead to inaccurate estimation \citep{sun2022reliable}. To illustrate, the current traffic assignment methodologies don't fully investigate the model performance under incomplete/missing OD demand information. This issue underscores a research gap in effectively solving TAPs and estimating traffic flows under inaccurate OD demand scenarios. 

Neural networks have demonstrated their capability for data imputation and data construction. Recently, convolutional neural networks (CNNs) \citep{fan2023deep} and graph neural networks (GNNs) \citep{rahman2023data} have been utilized to solve TAPs. Despite their improvement, these models encountered a few limitations. First of all, the CNN-based model cannot fully capture the topologies of transportation systems. Furthermore, the aforementioned models don't adequately consider the transportation network under various scenarios, e.g., link capacity reduction due to traffic accidents or lane closures due to maintenance. Last but not least, how these models perform under out-of-distribution data is not comprehensively explored. These limitations highlight the need for further research to enhance the adaptability and real-world applicability of neural networks for traffic assignment problems, specifically under the scenarios of unexpected events and incomplete OD demand.

It should be noted that the distribution of link flows  depends on the demand levels between different OD pairs. However, generally the graph representation of the transportation system only includes the physical roadway links, and  a direct link between  origin and destination nodes does not exist in the graph model for OD pairs that are not directly connected. This motivates us to  also include virtual links between  origin and destination node pairs in addition to the roadway links into the graph model. Considering this, we propose a novel heterogeneous GNN surrogate model, to also integrate comprehensive OD demand information and thereby  enhance  feature propagation across the network.  Furthermore, our proposed model includes a novel adaptive graph attention mechanism to propagate the node features efficiently,  a component to transform node embeddings into link embeddings for link flow and flow-capacity ratio estimation, and a  conservation-based loss function. To summarize, the major contributions of this work are as follows: (1) This is the first GNN learning of UE-TAP that integrates interdependencies between origin and destination nodes via a heterogeneous graph structure that consists of physical and virtual links and an adaptive attention-based mechanism; (2) the proposed model is trained using data and the governing conservation law and as a result, the estimated flows are more accurate; (3) due to the integration of virtual links and regularization based on the conservation law, the performance on unseen graphs is improved. In this paper, the efficiency and generalization capability of the proposed model are investigated through multiple experiments with different road network topologies, link characteristics, and OD demands.

The remainder of this article is structured as follows. Section \ref{sec:literature} provides a review of related literature. General backgrounds on the traffic assignment problem, the neural network and graph neural network models are presented in Section \ref{sec:background}. Section \ref{sec:architecture} includes the explanation of the proposed heterogeneous graph neural network for traffic assignment. Furthermore, the experiments with urban road networks and generalized synthetic networks are presented to demonstrate the accuracy and generalization capability of the proposed framework in Section \ref{sec:experiment}. Finally, the conclusion and discussion of the proposed framework are presented in Section \ref{sec:conslusion}.

\section{Related Literature}
\label{sec:literature}
The literature related to the estimation of traffic flows in transportation networks is discussed below under two areas of OD demand estimation  and traffic assignment.  By drawing insights from these related works, we have developed a robust and comprehensive approach to address the challenges associated with traffic assignment problems, particularly under network disruption events and incomplete OD demand data.

\subsection{OD Demand Estimation}

The primary challenge in predicting OD demands lies in the inability to directly measure demand using traffic sensors. Instead, researchers have developed models to infer the demand information  from aggregated data collected on roadways.  For instance, the autoregressive integrated moving average model has been developed to forecast traffic demand across different regions \citep{deng2011spatiotemporal}. Also, OD demands were estimated based on high-quality link flow counts using the Kalman filter  \citep{zhou2007structural}. In an optimization framework, \cite{zhang2020network}  formulated the OD estimation problem as a quadratic programming and then solved it using the alternating direction method of multipliers.

More recently, neural networks have been widely adopted for OD demand estimation due to their capability to model complex temporal and spatial dependencies in transportation datasets. For instance, \cite{xiong2020dynamic} integrated link graph neural networks with Kalman filters to predict OD demand. Also,  \cite{tang2021dynamic} employed a three-dimensional convolution neural network to learn the high-dimensional correlations between local traffic patterns and OD flows. Despite these methodological advancements, practical challenges such as sensor failures and malfunctions pose significant risks. These issues can result in the loss of relevant and reliable OD demand information, leading to inaccuracies in demand estimation, and consequently in the prediction of traffic flows \cite{sun2022reliable}.

\subsection{Traffic Assignment Problem}

Solving the traffic assignment 
problem provides a deeper understanding of traffic flow patterns and offers insights into managing traffic congestion. As a result, more efficient and reliable solution of traffic assignment problem has been the focus of many recent studies. For instance, \cite{lee2003conjugate} aimed at  improving convergence and computational efficiency in large-scale traffic networks by proposing a conjugate gradient projection  to enhance the gradient projection. Furthermore, \cite{babazadeh2020reduced} proposed a reduced gradient algorithm by selecting non-basic paths, to effectively manage computational complexity. However, these models rely on the assumption that the regional OD demand is accurate. In scenarios where OD demand information is incomplete or missing, conventional approaches tend to underestimate the link-wise traffic flow.

Recently, neural networks have demonstrated remarkable capabilities in data reconstruction and generalization \citep{zhang2018missing,liu2024neural}, presenting a potential solution for the issue of missing OD demands. However, the literature on neural networks in TAPs remains relatively scarce. One of the main approaches utilizes convolutional neural networks. For instance, \cite{fan2023deep} utilized recurrent CNNs for traffic assignment problems by considering the transportation network as the grid map, specifically under incomplete OD demands. Furthermore, since GNNs are inherently able to capture spatial information from graph topologies \citep{liu2022graph}, they were also used to solve TAPs \cite{rahman2023data}.

\subsection{Summary}

In summary, despite significant advances enabled by deep learning, a comprehensive graph neural network-based framework for equilibrium-based traffic assignment and link-wise flow estimation remains elusive. Firstly, it is challenging to recover the traffic flow distribution from incomplete OD demand. Secondly, the performance of these models under out-of-distribution data has not been thoroughly investigated. Last but not least, further investigations are required to adequately consider the transportation network under various scenarios, e.g., link capacity reduction and lane closures.

\section{Technical Background}
\label{sec:background}

\subsection{Traffic Assignment Problem}
The traffic assignment problem involves assigning traffic volumes or flows to each edge in the network. Given a transportation network represented as a graph $\mathcal{G} = (\mathcal{V}, \mathcal{E})$, where nodes $\mathcal{V}$ represent intersections of roads and edges $\mathcal{E}$ represent roads or links connecting these locations, The general form of traffic assignment problem can be considered as an optimization task:

\begin{equation}
    \label{eq:basic_formulation}
    \min_{f}:\quad \sum_{e\in \mathcal{E}} Z_e(f_e),
\end{equation}
where $f_e$ and $Z_e(f_e)$ are the total flow and the link cost function on link $e$, respectively. The link cost function can be expressed as the function of travel time, travel distance, or other relevant factors. Besides, the traffic assignment problem can have different formulations depending on the specific objectives and assumptions. For instance, the \citep{beckmann1956studies} addresses the user equilibrium traffic assignment problem by optimizing the following objective function:

\begin{equation}
\label{eq:UE-TAP}
\begin{aligned}
    \min:\quad & z(x) = \sum_{e\in\mathcal{E}} \int_0^{x_e} t_e(\omega) \mathrm{d}\omega \\
    \mathrm{s.t.}\quad & \sum_k f_k^{rs} = q_{rs},\  \forall r,s \in \mathcal{V}, \\
    & f_k^{rs} \geq 0,\  \forall k,r,s \in \mathcal{V}, \\
    & x_e = \sum_{rs}\sum_{k}f_k^{rs}\zeta^{rs}_{e,k}, \forall e \in \mathcal{E},
\end{aligned}
\end{equation}
where the objective function is the summation over all road segments of the integral of the link travel time function between 0 and the link flow. $t_e(\cdot)$ is the link travel time function, $q_{rs}$ is the total demand from source $r$ to destination $s$, $f^{rs}_k$ represents the flow on $k^{\mathrm{th}}$ path from $r$ to {s}. $\zeta^{rs}_{e,k}$ is the binary value, which equals 1 when link $e$ is on $k^{\mathrm{th}}$ connecting $r$ and $s$. It is noted that the objective function in Beckmann's formulation serves more as a mathematical construct for optimization than a direct physical representation. Compared with UE-TAP, SO-TAP changes the objective function to the summation of the travel time of all vehicles, which reflects a system-optimized perspective.

\subsection{Neural Networks}
\label{sec:nn}
Without loss of generality, we begin by considering a neural network with a single layer.  Given a $p$-dimensional input vector $\bm{h}^k \in \mathbb{R}^p$, $q$-dimensional the output $\bm{h}^{k+1} \in \mathbb{R}^q$ of single layer neural network, the layer indexed by $k$ can be expressed as:

\begin{equation}
\bm{h}^{k+1} = \sigma(\bm{h}^k \bm{W}_k + \bm{b}_k),
\end{equation}
where $\bm{W}_k \in \mathbb{R}^{p\times q}$ and $\bm{b}_k \in \mathbb{R}^{1\times q}$ represent the weight and bias term, respectively. The non-linear activation function $\sigma(\cdot)$ is utilized in the neural network. Theoretically, a single-layer neural network with an infinite number of neurons can approximate any continuous function to arbitrary accuracy, given a sufficiently large dataset \citep{hornik1989multilayer}. However, due to limitations in network width, dataset size, and the challenge of tuning parameters, a single-layer network is not optimal for achieving top performance, which leads to overfitting and poor generalization performance. To alleviate the limitation, multiple neural network layers are stacked together to enhance its expressibility and capture complex hierarchical features.

\subsection{Graph Neural Network}
\label{sec:gnn}
Neural networks have shown remarkable performance in various applications. In most neural network applications, input data structures are normally fixed, which is also called Euclidean data. However, non-Euclidean data structure such as graph-structured data is pervasive in different applications. The complexity and variability of the graph structure data make it difficult to model with conventional neural network architectures. To address this challenge, GNNs are specifically designed to handle graph-structured data. It operates on the node features and edge features and learns to extract embedding from nodes and edges, aiming to capture the underlying graph structure.

There are different types of graph neural network formulation. One of the popular approaches is the spectral approach \citep{wang2022powerful}. Spectral graph convolution is a type of convolution operation on graph signals that uses the graph Fourier transform. It operates in the frequency domain and utilizes the eigenvalues and eigenvectors of the graph Laplacian to filter the node features. Given a graph $\mathcal{G}=(\mathcal{V}, \mathcal{E})$ with adjacency matrix $\bm A$ and diagonal degree matrix $\bm D = \mathrm{diag}(\bm {A}\vec{\bm{1}})$, the Laplacian matrix and normalized Laplacian matrix of the graph is defined as $\bm{L} = \bm{D} - \bm{A}$ and $\bm{L}_{\mathrm{norm}} = \bm{D}^{-\frac{1}{2}} \bm{L} \bm{D}^{-\frac{1}{2}}$, respectively. The spectral graph convolution is defined mathematically as:

\begin{equation}
\label{eq:spectral}
    g_{\theta} \ast \bm{x} = \bm{U} g_{\theta}(\bm{U}^{T}\bm{x}),
\end{equation}
where $g_{\theta}$ is a filter with learnable parameters $\theta$, $\bm x \in \mathbb{R}^{|\mathcal{V}| \times N_F}$ is the input features with $|\mathcal{V}|$ nodes and $N_F$ features per node, and $\bm{U}$ is the eigenvectors of $\bm{L}_{\mathrm{norm}}$. The input signal is first transformed into the spectral domain. The features are passed through the learnable filter and transformed back into the spatial domain. The graph spectral operator can be applied to graphs of varying sizes. As a different approach to modeling graph data, the graph attention network (GAT) learns the graph feature by computing attention scores for each node based on its features and the features of its neighbors \citep{velivckovic2017graph}. The graph attention network computes the new node representation $\bm{x}_i'$ for each node $i$ as follows:

\begin{equation}
\label{eq:gat_score}
    \bm{x}'_i = \sigma\left(\sum_{j=1}^N \alpha_{ij} \bm{W}_x \bm{x}_j\right),
\end{equation}
where $\sigma$ is an activation function, $\bm{W}_x$ is a learnable weight matrix, and $\alpha_{ij}$ is the attention weight assigned to the node $j$ related to its neighbour node $i$. The attention weights are computed as follows:

\begin{equation}
\label{eq:gat_weight}
    \alpha_{ij} = \frac{\exp(\sigma(\bm{a}^T[\bm{W}_x \bm{x}_i \oplus \bm{W}_x \bm{x}_j]))}{\sum_{k\in \mathcal{N}(i)} \exp(\sigma(\bm{a}^T[\bm{W}_x \bm{x}_i \oplus \bm{W}_x \bm{x}_j]))}, 
\end{equation}
where $\bm{a}$ is a learnable weight vector, $\mathcal{N}(i)$ is the set of neighboring nodes of node $i$, $\oplus$ denotes concatenation function. The graph attention mechanism can be stacked into multiple layers, with each layer learning increasingly complex representations of the graph. The attention mechanism allows the network to learn the different importance of different nodes within a neighborhood, which can improve model performance.

The aforementioned formulation is valid for homogeneous graphs, where all nodes and edges have the same semantic meaning. However, it is noted that real-world graphs are not always homogeneous. For instance, in the literature citation graph, nodes can represent various entities such as papers, authors, and journals, while edges may denote different semantic relationships. When the graph contains different types of nodes or edges, it is considered as a heterogeneous graph. Utilizing GNNs on heterogeneous graphs offers notable advantages over homogeneous counterparts, particularly in the ability to learn type-specific representations for each node and edge type \citep{wang2019heterogeneous,fu2020magnn}. This allows for more accurate and targeted modeling of each entity and relationship, leading to improved performance on downstream tasks \citep{zhao2021heterogeneous}. In the following sections, we will leverage the expressiveness of the heterogeneous graph neural network to estimate the traffic flow performance under different OD demand settings.

\section{Heterogeneous Graph Neural Networks for Traffic Assignment}
\label{sec:architecture}
\begin{figure}[!hbt]
  \includegraphics[width=\linewidth]{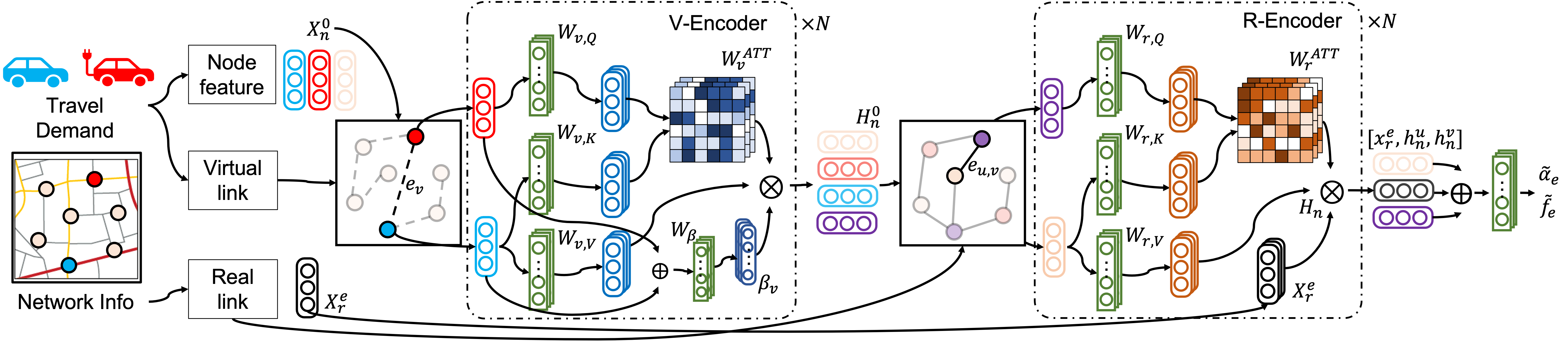}
  \caption{The illustration of the heterogeneous graph neural network for traffic assignment. The proposed model consists of three parts: graph construction \& feature preprocessing module; spatial feature extraction module, and edge prediction module. The graph features are first passed into the virtual encoder (V-Encoder) through the virtual links. Then the graph embeddings are passed into the real encoder (R-Encoder) through the real links. The flow-capacity ratio and link flow of each link are calculated using the source node feature, destination node feature, and normalized edge feature.}
  \label{fig:model}
\end{figure}

In this section, we elaborate on the proposed architecture of the heterogeneous graph neural networks for traffic assignment. The illustration of the proposed model is shown in Fig. \ref{fig:model}. It consists of three modules: graph construction \& feature preprocessing module; spatial feature extraction module, and edge prediction module. The detail explanation of each module is described as follows.

\subsection{Graph Construction \& Feature Preprocessing}
The heterogeneous graph $\mathcal{G} = (\mathcal{V}, \mathcal{E}_r, \mathcal{E}_v)$ for traffic assignment consists of one type of node representing the intersections of road segments and two edge types: real links and virtual links. The real links represent the road segments in the road network, while the virtual links represent the auxiliary link between the origin and destination nodes. The incorporation of the virtual link, as an edge augmentation technique, is strategically employed to facilitate enhanced feature updating. In this model, the node feature attribute is denoted as $\bm{X}_n \in \mathbb{R}^{|\mathcal{V}| \times (|\mathcal{V}|+2)}$, where $\mathcal{V}$ represents the graph's node set. Specifically, each row $\bm{x}_u \in \mathbb{R}^{1 \times (|\mathcal{V}|+2)}$ within this matrix corresponds to the feature vector of a single node $u \in \mathcal{V}$ including the origin-destination demand as well as geographical coordinates. The feature representation for the entire set of real edges is represented with $\bm{X}_{e,r} \in \mathbb{R}^{|\mathcal{E}_r| \times 2}$. Each row in this matrix includes the free-flow travel time and the link capacity for a single link. 

Furthermore, the original node features are often sparse and non-normalized. To address this issue, we employ a preprocessing step to encode the raw features into a lower-dimensional representation. This process not only captures the essential attributes of the data but also retains the semantic information. The generated node feature embedding size is $\bm{X}^0_n \in \mathbb{R}^{|\mathcal{V}| \times N_v}$, where $N_v$ is the embedding size. Similarly, the edge features are also normalized before being propagated in the message passing. Additionally, it should be noted that there is an overlap between the real links and virtual links in the heterogeneous graph because it is possible that a road and an OD demand both exist between the same node pair.

\subsection{Graph Spatial Features Extraction}
\label{sec:extracting}
As discussed in Section \ref{sec:gnn}, the key challenge in heterogeneous graphs lies in effectively aggregating diverse node and edge information. To address this challenge, we propose a novel sequential graph encoder for feature extraction and propagation. The sequential graph encoder is twofold: virtual graph encoder (V-Encoder) and real graph encoder (R-Encoder). Each part leverages attention mechanisms tailored to either virtual or real links, respectively. We will elaborate on these components in the following sections.

\subsubsection{Virtual Graph Encoder (V-Encoder)}
As the first step of spatial feature extraction, we employ a graph transformer-based attention mechanism on graph features to enhance the modeling capability. Specifically, we transform node features into key, query, and value matrices and then calculate the attention score between node pairs. Furthermore, since the virtual links are synthetically generated without inherent edge features, we introduce a learnable adaptive weight for virtual links, serving as their edge features. The adaptive weight for each node pair is derived by concatenating the two node features and passing through a position-wise feed-forward network (FFN). Mathematically, $L^{\mathrm{th}}$ V-Encoder can be expressed as:

\begin{equation}
\begin{aligned}
\label{eq:virtual_attention_part1}
\bm{q}^{L,i}, \bm{k}^{L,i}, \bm{v}^{L,i} & = \bm{x}^{L}_{u} \ [\bm{W}_{q}^{L,i}, \bm{W}_{k}^{L,i}, \bm{W}_{v}^{L,i}], & \forall u \in \mathcal{V}\\
\beta_{e}^{L,i} & = \mathtt{FFN}\left([\bm{x}^{L,i}_{u} \oplus \bm{x}^{L,i}_{v}]; \bm{W}^{L,i}_{\beta}, \bm{b}^{L,i}_{\beta}
\right), & \forall e=(u,v) \in \mathcal{E}_v\\
s_{e}^{L,i} &= \mathrm{exp} \left(\frac{\bm{q}_u^{L,i} \bm{k}_v^{L,i}}{\sqrt{d_L}} \beta_{e}^{L,i} \right), & \forall e=(u,v) \in \mathcal{E}_v
\end{aligned}
\end{equation}
where $\bm{x}^{L}_{u}$ is the feature embedding of node $u \in \mathcal{V}$ at $L^{\mathrm{th}}$ layer of V-Encoder. $\bm{q}_{(\cdot)} \in \mathbb{R}^{d_L}$, $\bm{k}_{(\cdot)} \in \mathbb{R}^{d_L}$, and $\bm{v}_{(\cdot)} \in \mathbb{R}^{d_L}$ is the query, key, and value vector at $i^{\mathrm{th}}$ head in V-Encoder where $d_L$ denotes the dimensionality of the feature vectors. $\bm{W}_{(\cdot)}$ and $\bm{b}_{(\cdot)}$ is the learnable parameters. $\beta_{e}^{L,i} \in \mathbb{R}$ and $s_{e}^{L,i} \in \mathbb{R}$ represents the learnable adaptive weight and the unnormalized attention score of the edge $e=(u, v) \in \mathcal{E}_v$ at $i^{\mathrm{th}}$ head of $L^{\mathrm{th}}$ layer, respectively. $\oplus$ is the concatenation operator. The introduction of edge-level adaptive weights is motivated by the fact that the observed variability in low-dimensional node embedding, in turn reflects the variability in node OD demand. Specifically, node pairs with higher OD demands should receive higher attention scores since they have more significant impacts on the flow distribution. By introducing the adaptive edge-level weight, the graph encoder can adjust the attention score among these node pairs and adaptively propagate the most relevant information through virtual links, thereby enriching the model's contextual understanding. Then value vectors normalized by the attention scores, are processed through another position-wise feed-forward network, accompanied by layer normalization. Additionally, a residual connection supplements the final output of the layer, ensuring the integration of original input features with learned representations for enhanced model performance. Finally, the output of the V-encoder is obtained by concatenating the outputs from all attention heads:

\begin{equation}
\label{eq:virtual_attention}
\begin{aligned}
\bm{z}^{L,i}_u &= \sum_{v \in \mathcal{N}_o(u)} s_{(u,v)}^{L,i} \bm{v}_v^{L,i} / \sum_{v \in \mathcal{N}_o(u)} s_{(u,v)}^{L,i}, & \\
\bm{x}_u^{L+1,i} & = \bm{x}_u^{L,i} + \mathtt{LayerNorm} \left(\mathtt{FFN} \left(\bm{z}_u^{L,i}; \bm{W}^{L,i}_{z}, \bm{b}^{L,i}_{z}\right)\right), & \forall u \in \mathcal{V}\\
\bm{x}_u^{L+1} & = \left[\bm{x}_u^{L+1,0} \oplus \bm{x}_u^{L+1,1}\oplus \dots \oplus \bm{x}_u^{L+1,N_h} \right], &
\end{aligned}
\end{equation}
where $\bm{z}^{L,i}_u$ represents the normalized weighted vector of node $u$ and $\mathcal{N}_o(u)$ represents all the outgoing nodes connected to $u$. $N_h$ represents the number of attention heads of the V-Encoder. To enhance the propagation of node features throughout the network, we sequentially stack multiple layers of the V-Encoder. The final output of this stacked V-Encoder, denoted as $H_n^0$, serves as the input for the subsequent encoder.

\subsubsection{Real Graph Encoder (R-Encoder)}
The R-Encoder is designed to enhance and complement the functionality of the V-Encoder. These two encoders share a similar architecture but slightly differ in the graph attention score mechanism. To illustrate, some nodes in the V-Encoder do not exchange messages with others due to the absence of virtual link connections. The R-Encoder addresses this issue by updating node features through real links, ensuring a comprehensive assessment of direct and indirect node relationships. From a mathematical perspective, the $M^{\mathrm{th}}$ layer of R-Encoder is expressed as follows:

\begin{equation}
\label{eq:real_attention}
\begin{aligned}
\bm{q}^{M,j}, \bm{k}^{M,j}, \bm{v}^{M,j} & = \bm{h}^{M}_{u} \ [\bm{W}_{q}^{M,j}, \bm{W}_{k}^{M,j}, \bm{W}_{v}^{M,j}], & \forall u \in \mathcal{V}\\
s_{e}^{M,j} &= \mathrm{exp} \left( \sum_{p=1}^{P} \frac{\bm{q}_u^{M,j} \bm{k}_v^{M,j}}{\sqrt{d_L}}\beta^r_{e,p} \right), & \forall e=(u,v) \in \mathcal{E}_r \\
\bm{z}^{M,j}_u &= \sum_{v \in \mathcal{N}_o(u)} s_{(u,v)}^{M,j} \bm{v}_v^{M,j} / \sum_{v \in \mathcal{N}_o(u)} s_{(u,v)}^{M,j}, & \forall u \in \mathcal{V} \\
\bm{h}_u^{M+1,j} & = \bm{h}_u^{M,j} + \mathtt{LayerNorm} \left(\mathtt{FFN} \left(\bm{z}_u^{M,j}; \bm{W}^{M,j}_{z}, \bm{b}^{M,j}_{z}\right)\right), & \forall u \in \mathcal{V}\\
\bm{h}_u^{M+1} & = \left[\bm{h}_u^{M+1,0} \oplus \bm{h}_u^{M+1,1}\oplus \dots \oplus \bm{h}_u^{M+1,N_h} \right], & \forall u \in \mathcal{V}
\end{aligned}
\end{equation}
where $\bm{h}^{M}_{u}$ is the feature embedding of node $u \in \mathcal{V}$ at $M^{\mathrm{th}}$ layer of R-Encoder. $\bm{q}_{(\cdot)} \in \mathbb{R}^{d_L}$, $\bm{k}_{(\cdot)} \in \mathbb{R}^{d_L}$, and $\bm{v}_{(\cdot)} \in \mathbb{R}^{d_L}$ is the query, key, and value matrices at $j^{\mathrm{th}}$ head of R-Encoder. $\bm{W}_{(\cdot)}$ and $\bm{b}_{(\cdot)}$ are the learnable parameters. $\beta^r_{e,p}$ represents the $p^{\mathrm{th}}$ normalized edge feature of link $e \in \mathcal{E}_r$. Despite the subtle difference in the attention mechanism, the node features are propagated in two distinct patterns in the V-Encoder and R-Encoder. To illustrate, the V-Encoder captures the long-range dependency between nodes and integrates the contextual information from non-adjacent nodes. On the contrary, the R-Encoder captures the local topological relationships. Compared with the homogeneous graph with only real links, the additional feature propagation through virtual links can be considered as a dimension-reduction technique to reduce the number of hops required for distant nodes to gather messages. Consequently, it requires fewer GNN layers for effective feature aggregation and following edge prediction. Similar to the V-Encoder, the multiple R-Encoder layers are stacked together, and the output of the last R-Encoder $\bm{O}$, serves as the input for link flow prediction. 
% The node features passed along the virtual link can provide additional information for the following edge prediction.

\subsection{Graph Edge Prediction}
To predict the traffic flow at the edge level, the node embedding of the source node and destination node, and the normalized real edge feature are concatenated and passed through a feed-forward neural network. In this paper, we consider the flow-capacity ratio $\tilde{\alpha}_e$ as the quantity of the final link prediction, which is the link flow normalized by the link capacity:

\begin{equation}
\label{eq:edge_prediction}
\tilde{\alpha}_e = \texttt{MLP}([\bm{o}_{u} \oplus \bm{o}_{v} \oplus \bm{\beta}^r_{e}]; \bm{W}_o, \bm{b}_o),\quad \forall e=(u,v) \in \mathcal{E}_r
\end{equation}
where $\bm{o}_{(\cdot)}$ represent the node embedding. $\bm{W}_o$ and $\bm{b}_o$ is the learnable parameters associated with the multilayer perception. The predicted link flows $\tilde{f}_e$ can be calculated by multiplying the link capacity with the predicted flow-capacity ratio. Subsequently, selecting an appropriate loss function becomes crucial to ensure the model's effective convergence. The proposed model employs a composite loss function comprising two components. The first part is the supervised loss, which measures the difference between prediction and ground truth. It considers both the discrepancy from the flow-capacity ratio $L_{\alpha}$ and the link flow $L_{f}$ on each link:
\begin{equation}
\begin{split}
L_s & = L_{\alpha} + L_{f} \\
    & = \frac{1}{|\mathcal{E}_r|} \sum_{e \in \mathcal{E}_r}\|{\alpha}_e - \tilde{\alpha}_e \| + \frac{1}{|\mathcal{E}_r|} \sum_{e \in \mathcal{E}_r}\|f_e - \tilde{f}_e \| , 
    \label{eq:ls}
\end{split}
\end{equation}
where the ${\alpha}_e$ and $\tilde{\alpha}_e$ represent the ground truth and prediction of flow-capacity ratio on link $e \in \mathcal{E}$. The $f_e$ and $\tilde{f_e}$ represent the ground truth and prediction of link flow on link $e \in \mathcal{E}$. The second part of the loss function originates from the principle of node-based flow conservation, where the total flow of traffic entering a node equals the total flow of traffic exiting that node. The node-based flow conservation law can be represented mathematically:
%, as shown in Figure \ref{fig:conservation}

\begin{equation}
    \sum_k f_{ki} - \sum_j f_{ij} = \Delta f_i = \begin{cases}
 \quad \sum_{v\in\mathcal{V}} O_{v,i} - \sum_{v\in\mathcal{V}} O_{i,v}, & \text{ if } i \in \mathcal{V}_{OD}, \\ 
 \quad 0 & \text{ otherwise }, 
\end{cases}
\end{equation}
where $f_{ki}$ denotes the flow on the link $(k, i)$, $\Delta f_i$ represents the difference between flow receiving and sending at node $i$, $O_{v, i}$ represents the number of OD demand from $v$ to $i$. $\mathcal{V}_{OD}$ denote the origin-destination node set. The node-based flow conservation law can be considered as a normalization loss, thereby ensuring compliance with the fundamental principle of flow conservation. One common way to incorporate conservation law into the loss function is to define a residual loss function:

\begin{equation}
    L_c = \sum_{i} \ |\sum_{k \in \mathcal{N}_{i}(i)} \tilde{f}_{ki} - \sum_{j \in \mathcal{N}_{o}(i)} \tilde{f}_{ij} - \Delta f_i |,
    \label{eq:lf}
\end{equation}
where $\mathcal{N}_{i}(i)$ represent the incoming edges of node $i$. The normalization loss $L_{c}$ measures how the flow prediction satisfies the flow conservation law. Minimizing this loss function during training will encourage the model to learn traffic flow patterns that satisfy the conservation law. Consequently. the total loss for the flow prediction $L_{total}$ is the weighted summation of the supervised loss and the conservation loss:

\begin{equation}
    L_{total} = w_{\alpha} L_{\alpha} + w_{f} L_{f} + w_c L_c,
    \label{eq:total_loss}
\end{equation}
where the $w_{\alpha}$, $w_f$ and $w_c$ represent the normalized weight for supervised loss of flow-capacity ratio, supervised loss of actual flow, and the conservation loss, respectively. 

\section{Numerical Experiments}
\label{sec:experiment}
Two numerical experiments are conducted to evaluate the accuracy, efficiency, and generalization capability of the proposed graph neural network. The first experiment is on urban transportation networks. The second experiment is on multiple synthetic graphs with different topologies. The details of the experiments will be explained in the following sections.

\begin{table}[!htb]
\renewcommand{\arraystretch}{1.15}
\centering
\caption{The detail of urban transportation network. Three networks, including Sioux Falls, East Massachusetts, and Anaheim, are considered.}
\label{tab:urban_network_detail}
\resizebox{0.5\textwidth}{!}{%
\begin{tabular}{ccccc}
\hline\hline
Network Name & $|\mathcal{V}|$ & $|\mathcal{E}|$ & Average Degree & OD Demand \\ \hline
Sioux Falls  & 24      & 76      & 3.17           & 188,960          \\
EMA          & 74      & 258     & 3.49           & 132,106          \\
Anaheim      & 416     & 914    & 3.05           & 226,279          \\ \hline\hline
\end{tabular}%
}
\end{table}

\subsection{Experiments on Urban Transportation Networks}
\label{sec:urban}

\subsubsection{Characteristics of networks}
\begin{figure}[!htb]
\centering
\begin{subfigure}[b]{0.33\textwidth}
    \centering
    \includegraphics[width=\textwidth]{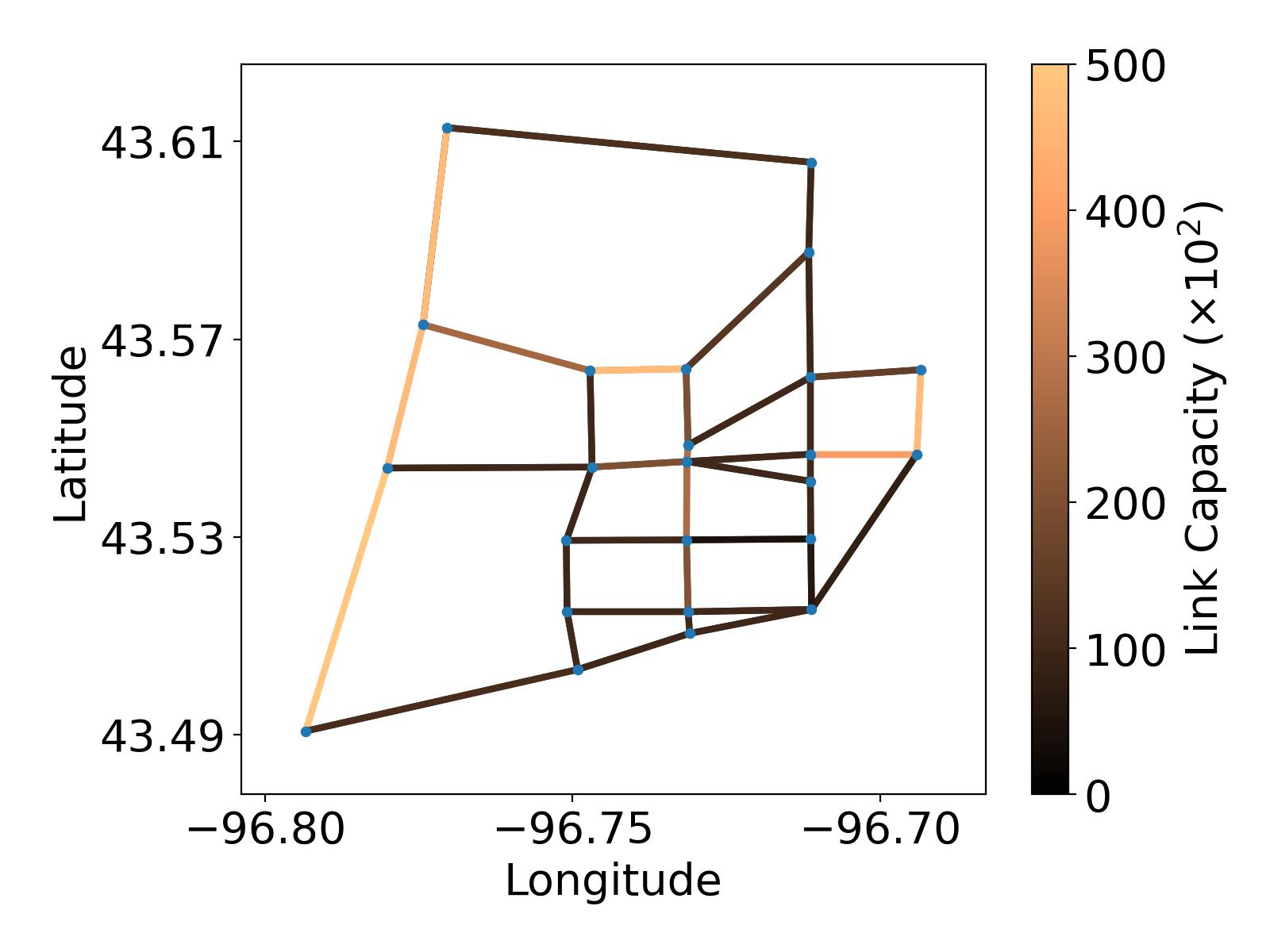}
    \caption{Sioux Falls network}
    \label{fig:Sioux}
\end{subfigure}
\hfill
\begin{subfigure}[b]{0.32\textwidth}
    \centering
    \includegraphics[width=\textwidth]{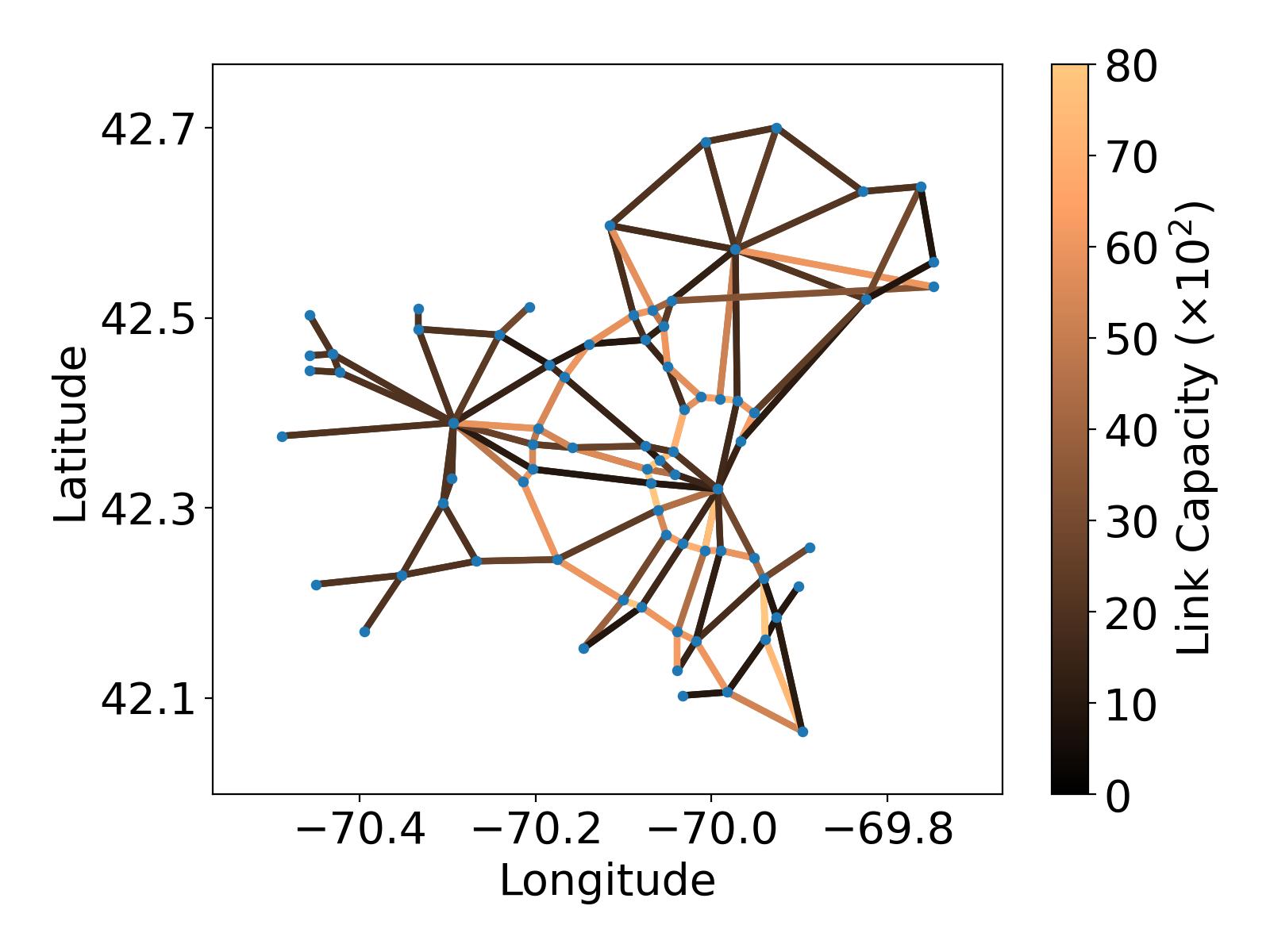}
    \caption{EMA network}
    \label{fig:EMA}
\end{subfigure}
\hfill
\begin{subfigure}[b]{0.33\textwidth}
    \centering
    \includegraphics[width=\textwidth]{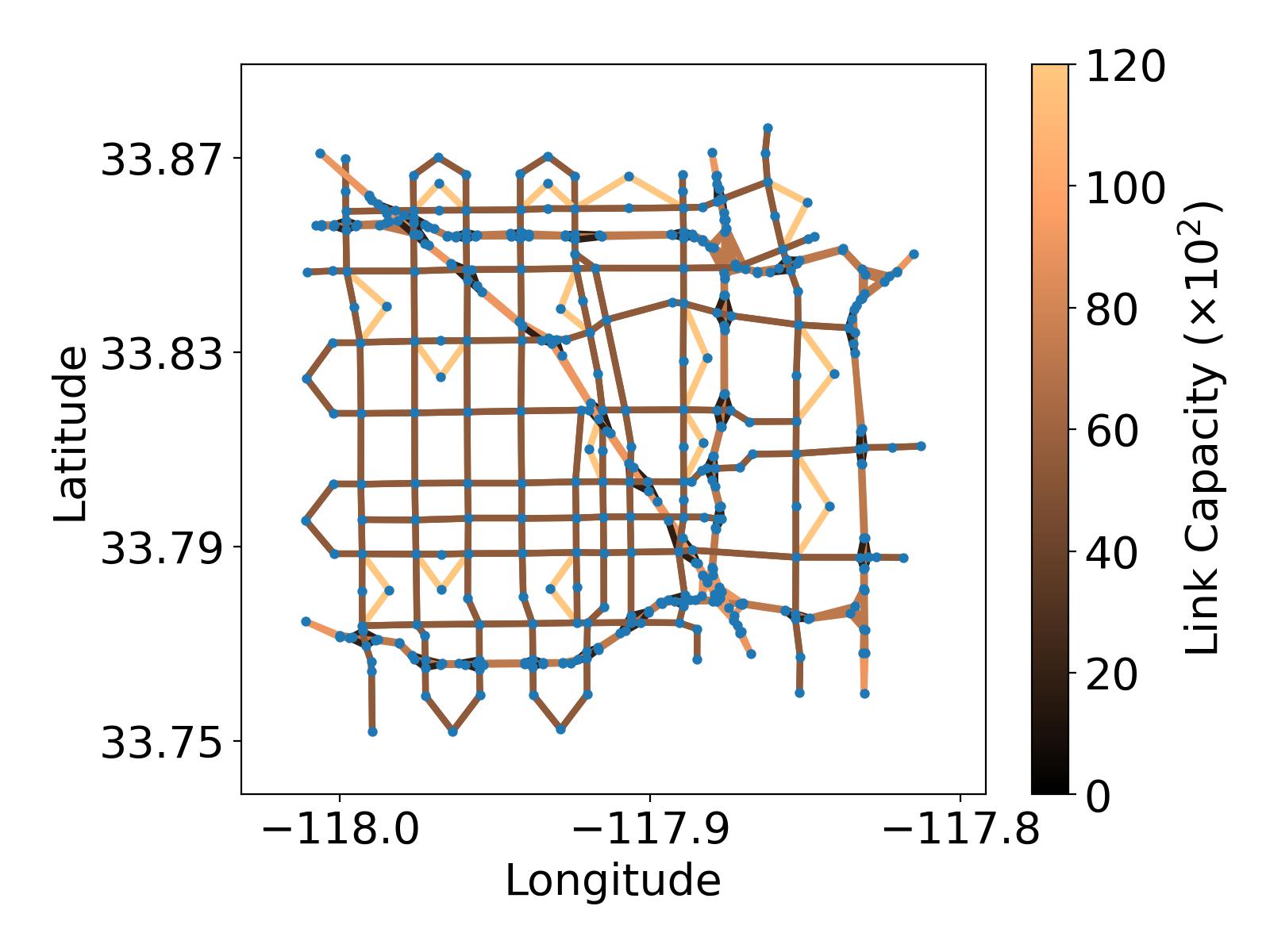}
    \caption{Anaheim network}
    \label{fig:anaheim}
\end{subfigure}
    \caption{The illustrations of urban transportation networks, including Sioux Falls, EMA, and Anaheim.}
    \label{fig:urban_network}
\end{figure}

As case studies, three urban transportation networks are selected: Sioux Falls network, East Massachusetts Network (EMA), and Anaheim network. The information about the network topology, link characteristics, and the OD demand of these networks are obtained from \citep{bar2021transportation}. The statistics and the illustration of the network topologies are shown in Table \ref{tab:urban_network_detail} and Figure \ref{fig:urban_network}, respectively. To create demand variation, we scaled the demand by a scaling factor according to
\begin{equation}
\label{eq:od_demand}
    \tilde{O}_{s,t} = \delta^o_{s,t} \ O_{s,t},
\end{equation}
where $O_{s,t}$ is the default OD demand between source $s$ and destination $t$ and $\delta^o_{s,t} \sim U(0.5, 1.5)$ is the uniformly distributed random scaling factor for the OD pair ($s$, $t$). Additionally, to account for variations in network properties, variable link capacities are created according to 
\begin{equation}
\label{eq:ca_demand}
    \tilde{c}_{a} = \delta^c_{a} \ c_{a} ,
\end{equation}
where $c_a$ is the original link capacity for link $a$, and $\delta^c_{a}$ is the scaling factor for link $a$. Capacity variations are considered to be due to traffic accidents, road construction/damage, and adverse weather conditions, which reduce the link capacity. In this work, three levels of capacity reduction are considered: (L): light disruption with $\delta^c_{a} \sim U(0.8, 1.0)$; (M) moderate disruption with $\delta^c_{a} \sim U(0.5, 1.0)$; (H) high disruption with $\delta^c_{a} \sim U(0.2, 1.0)$. 

\subsubsection{Training setup and model architecture}
\begin{figure}[!htb]
\centering
\begin{subfigure}[b]{0.3\textwidth}
    \centering
    \includegraphics[width=\textwidth]{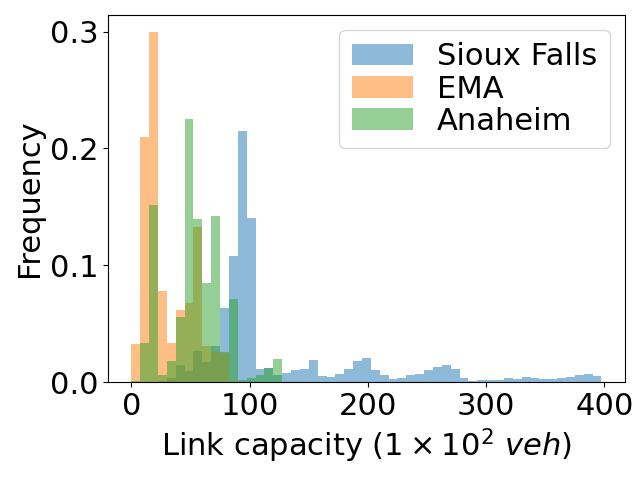}
    \label{fig:all_link_capacity_variation}
\end{subfigure}
\quad\quad
\begin{subfigure}[b]{0.3\textwidth}
    \centering
    \includegraphics[width=\textwidth]{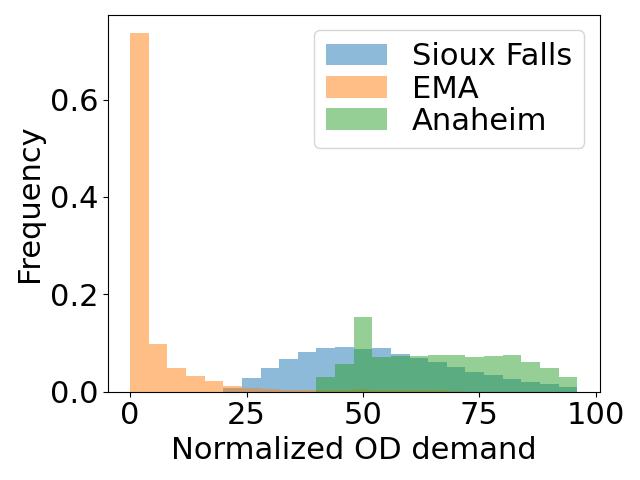}
    \label{fig:all_od_variation}
\end{subfigure}
\caption{The histogram of the link capacity and OD demand in the training and testing data. Three transportation networks are considered including Sioux Falls, EMA, and Anaheim network.}
\label{fig:variation_histogram}
\end{figure}

The size of the dataset for each network at each disruption scenario is 5000, which is split into the training set and the testing set with a ratio of 80\% and 20\%, respectively. To demonstrate the dataset is sufficiently diverse to cover enough scenarios, the coefficient of variation of network link capacity and OD demand is calculated and the histogram of the link capacity and the OD demand of training and testing data is shown in Figure \ref{fig:variation_histogram}. The OD demand of each network is normalized to 100 in order to facilitate a standardized comparison across different networks regardless of their actual size or demand volumes. The minimum coefficient of variation of link capacity and OD demand among the three networks is 0.45 and 0.22, respectively, which indicates the training and testing data are sufficiently diverse to cover different scenarios \citep{bedeian2000use, campbell2010medical}.

The training and testing dataset are obtained by solving UE-TAP with the Frank-Wolfe algorithm \citep{fukushima1984modified}. The algorithm converges when the square root of the sum of the squared differences between the link flow in two successive iterations, normalized by the sum of the values of the link flow, falls below the threshold of 1e-5. The GNN model is implemented using PyTorch \citep{paszke2019pytorch} and DGL \citep{wang2019deep}. The preprocessing layer consists of a three-layer fully connected neural network with an embedding size of 32. The number of GNN layers in the proposed model is 4, including two V-Encoders and two R-Encoders. The number of heads in the attention block is 8. For hyper-parameter selection, the hidden layer size is chosen as 64, which is common in neural network implementation \citep{liu2023physics}. The learning rate and batch size of training are 0.001 and 128, respectively. The weights of $L_\alpha$, $L_f$, and $L_c$ in equation \ref{eq:total_loss} are chosen as 1.0, 0.005, and 0.05, respectively.

We evaluated the performance of our proposed heterogeneous GNN model (referred to by HetGAT) and compared it with three benchmark models: a fully connected neural network (FCNN), a homogeneous graph attention network (GAT), and a homogeneous graph convolution network (GCN) \citep{rahman2023data}. The FCNN consisted of five fully connected layers with an embedding size of 64. The GAT and GCN both have four layers of graph message passing layer, followed by three layers of FCNN with an embedding size of 64. The metrics to evaluate performance include the mean absolute error (MAE), root mean square error (RMSE), and the normalized conservation loss $\tilde{L}_c$:

\begin{equation}
\mathrm{MAE} = \frac{1}{N}\sum_{i=1}^{N}|y_i-\tilde{y}_i|,
\end{equation}

\begin{equation}
\mathrm{RMSE} = \sqrt{\frac{1}{N}\sum_{i=1}^{N}(y_i-\tilde{y}_i)^2},
\end{equation}

\begin{equation}
\tilde{L}_c = \frac{\sum_{i} | \sum_{k \in \mathcal{N}_{i}(i)} \tilde{f}_{ki} - \sum_{j \in \mathcal{N}_{o}(i)} \tilde{f}_{ij} - \Delta f_i |}{\sum_{s}\sum_{t}\tilde{O}_{s,t}},
\end{equation}
where $y$ and $\tilde{y}$ respectively represent the ground truth and predicted values for quantity of interest. We conducted a 5-fold cross-validation for each experiment to ensure the robustness of our results across different subsets of the data.

\subsubsection{Numerical results}
\begin{figure}[!htb]
\centering
\begin{subfigure}[b]{0.33\textwidth}
    \centering
    \includegraphics[width=\textwidth]{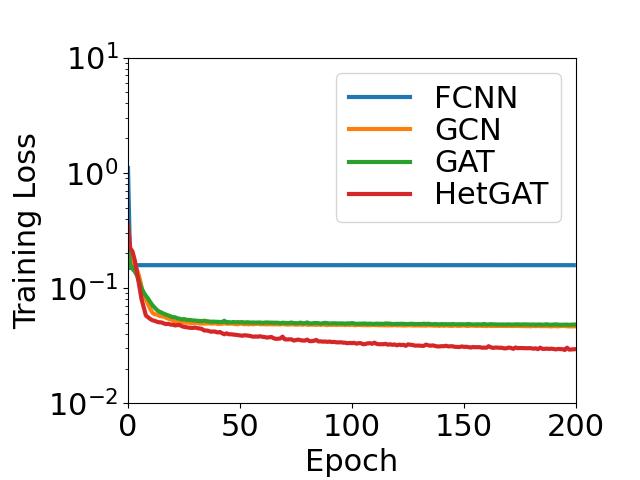}
    \caption{Sioux Falls network}
    \label{fig:Sioux_loss}
\end{subfigure}
\hfill
\begin{subfigure}[b]{0.32\textwidth}
    \centering
    \includegraphics[width=\textwidth]{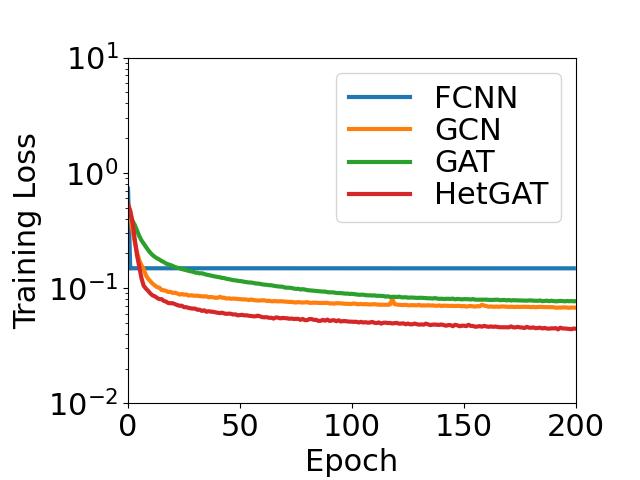}
    \caption{EMA network}
    \label{fig:EMA_loss}
\end{subfigure}
\hfill
\begin{subfigure}[b]{0.33\textwidth}
    \centering
    \includegraphics[width=\textwidth]{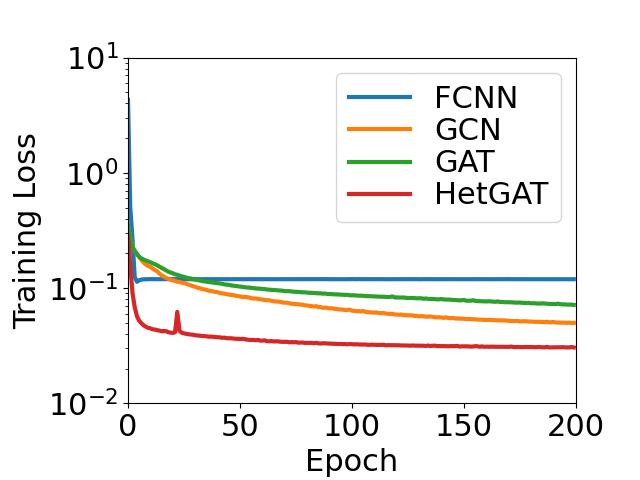}
    \caption{Anaheim network}
    \label{fig:anaheim_loss}
\end{subfigure}
    \caption{Training loss history under urban transportation network. Three benchmarks, including FCNN, GCN, and GAT, are compared with HetGAT.}
    \label{fig:anaheim_all_loss}
\end{figure}

The training histories of the studied models are shown in Figure \ref{fig:anaheim_all_loss}. The results indicate that the FCNN model performed poorly during training compared to GNN-based models, as shown by the high training loss and early stagnation. In contrast, GCN and GAT models exhibited similar convergence rates. Our proposed model outperformed both GCN and GAT in terms of training loss. Especially when used for larger networks, the proposed model demonstrated superior convergence performance compared to GCN and GAT; for the Anaheim network, the training loss of the proposed model is almost 1/3 of that of GAT in the first 25 iterations. This is because GCN and GAT only consider homogeneous edges, which limits the message passing to adjacent nodes. In contrast, the proposed GNN model uses virtual links and provides augmented connectivity to long-hop node pairs, which makes the node feature updating in HetGAT more efficient.

\begin{figure}[!htb]
\centering
\begin{subfigure}[b]{0.32\textwidth}
    \centering
    \includegraphics[width=\textwidth]{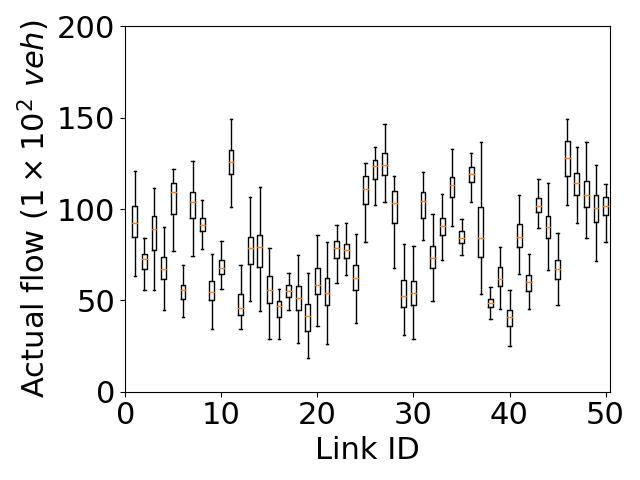}
    \caption{Sioux Falls network}
    \label{fig:flow_boxplot_Sioux}
\end{subfigure}
\begin{subfigure}[b]{0.32\textwidth}
    \centering
    \includegraphics[width=\textwidth]{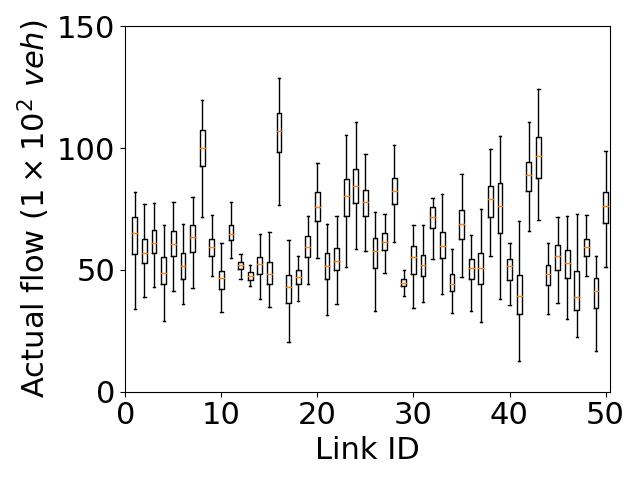}
    \caption{EMA network}
    \label{fig:flow_boxplot_EMA}
\end{subfigure}
\begin{subfigure}[b]{0.32\textwidth}
    \centering
    \includegraphics[width=\textwidth]{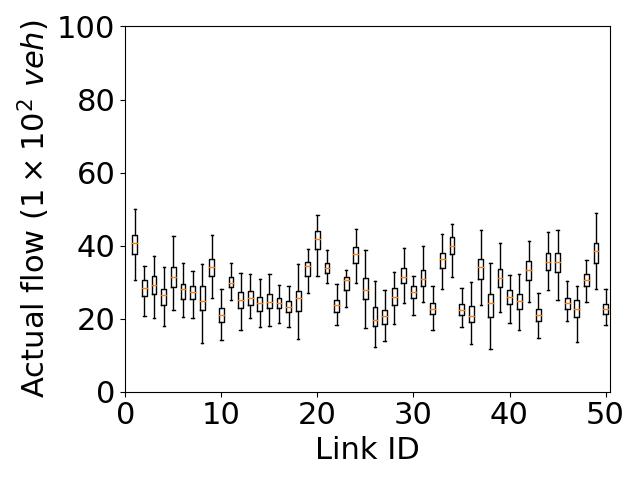}
    \caption{Anaheim network}
    \label{fig:flow_boxplot_anaheim}
\end{subfigure}
\caption{Illustrations of the link-wise flow distribution for different transportation networks. Three transportation networks under major disruption are considered, including Sioux Falls, EMA, and Anaheim networks. 50 links are selected for each network.}
\label{fig:flow_boxplot}
\end{figure}

\begin{figure}[htb!]
\centering
\begin{subfigure}[b]{0.26\textwidth}
    \centering
    \includegraphics[width=\textwidth]{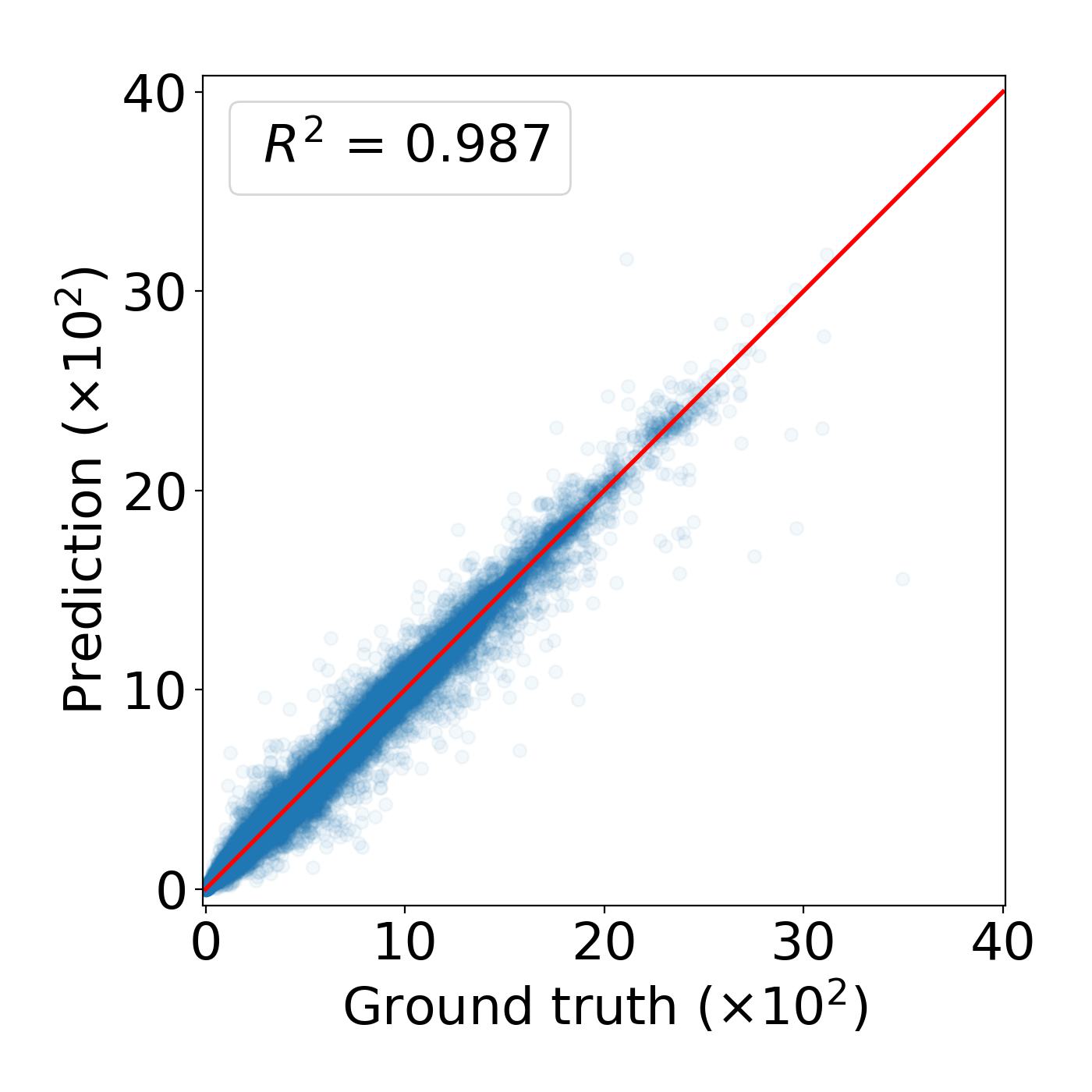}
    \caption{HetGAT}
    \label{fig:anaheim_HetGAT_flow_loss}
\end{subfigure}
\hfill
\begin{subfigure}[b]{0.26\textwidth}
    \centering
    \includegraphics[width=\textwidth]{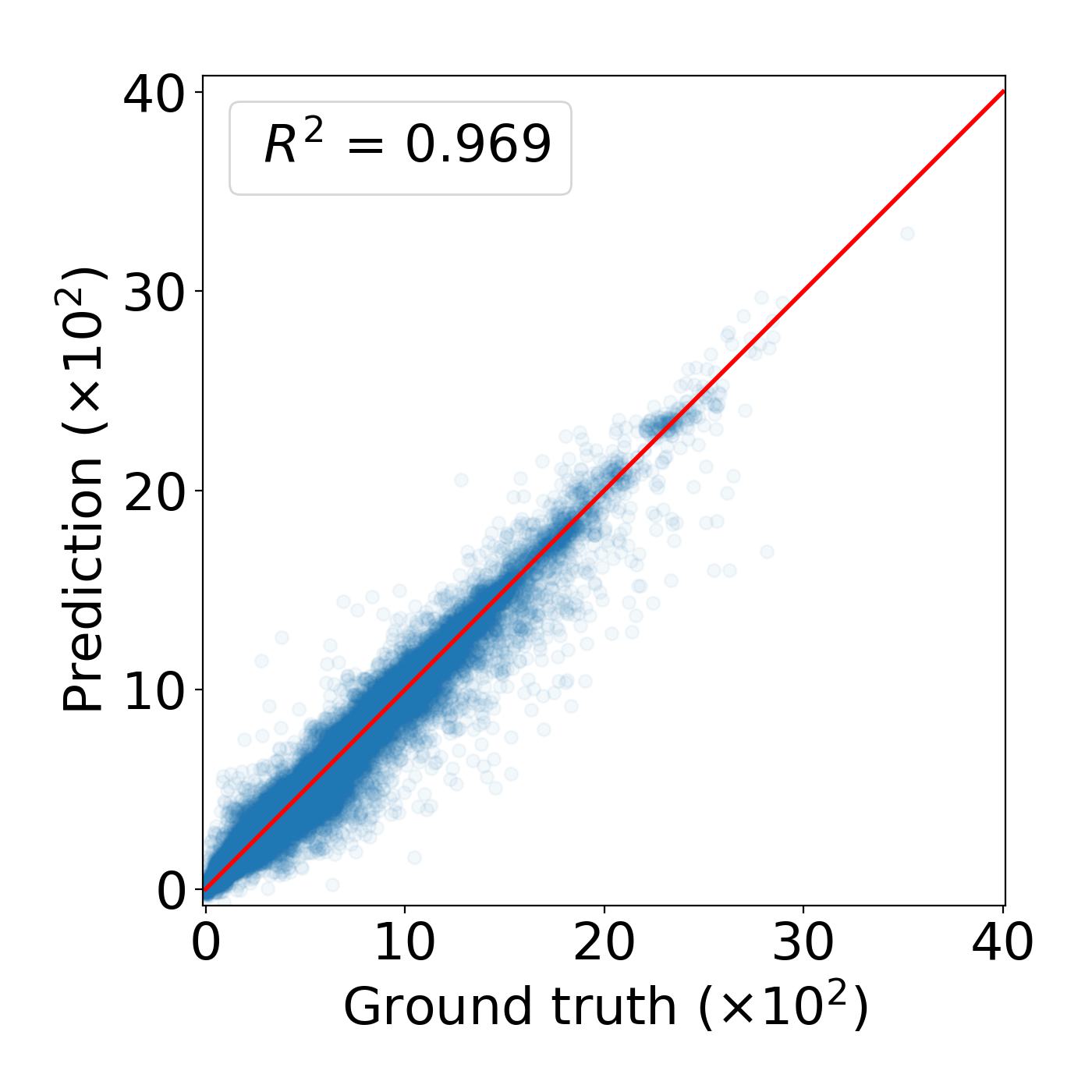}
    \caption{GAT}
    \label{fig:anaheim_GAT_flow_loss}
\end{subfigure}
\hfill
\begin{subfigure}[b]{0.26\textwidth}
    \centering
    \includegraphics[width=\textwidth]{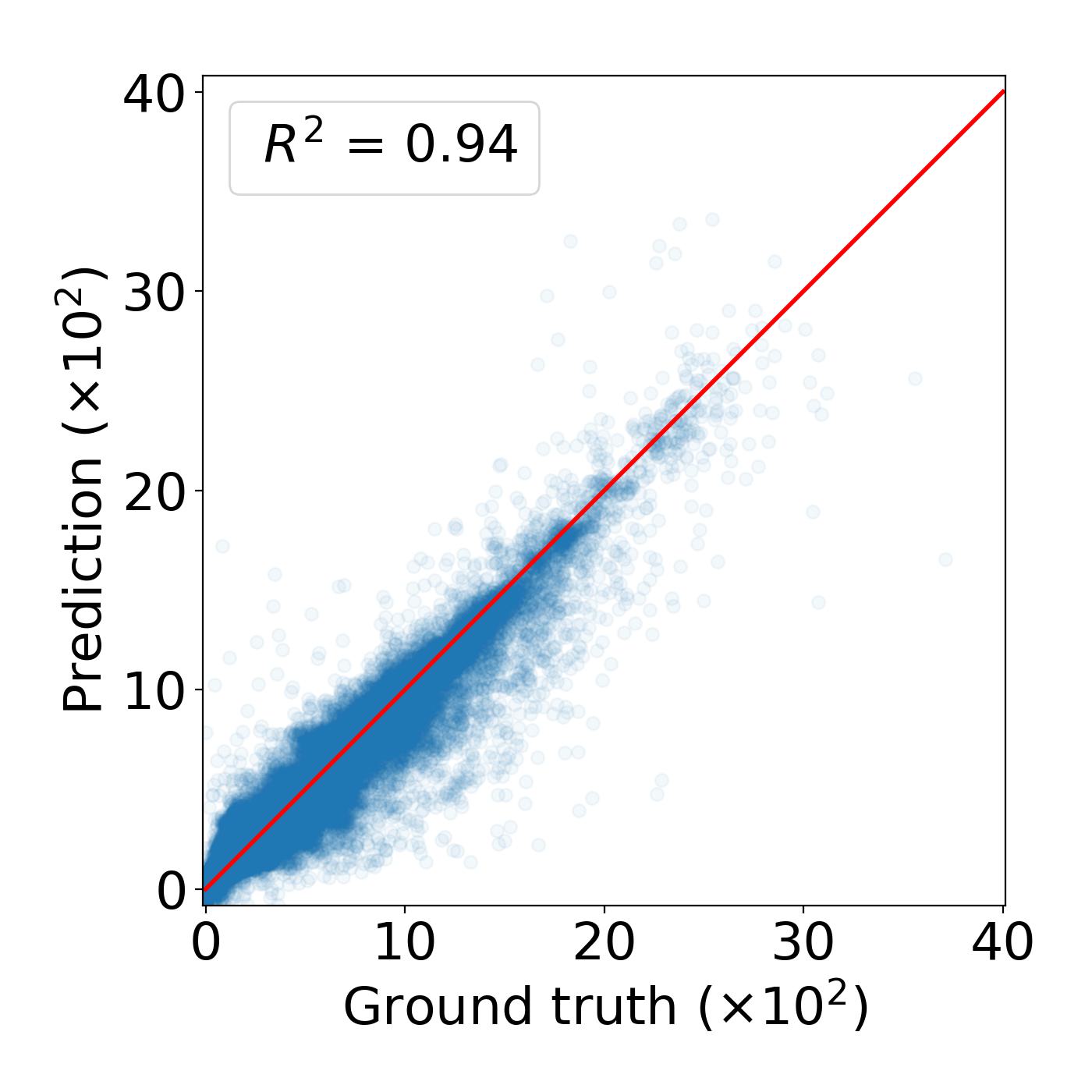}
    \caption{GCN}
    \label{fig:anaheim_GCN_flow_loss}
\end{subfigure}
    \caption{Comparison of link flow between ground truth and surrogate model prediction in the Anaheim network under LMH-LMH setting.}
    \label{fig:urban_flow_loss}
\end{figure}

After the training is finished, the model performance is evaluated on the testing set. The experiments under the urban road network are conducted in three different settings. The first setting, referred to as LMH-LMH, involved using all levels of disruption (flow reduction scaling levels) in both the training and testing sets. The second setting, namely L-M, involves training the model on light disruption data and testing it on medium disruption data. The third case, which is labeled as M-H, involves training on medium disruption data and testing on high disruption data.  The L-M  and M-H scenarios will therefore involve unseen cases that don't exist in the training data. Figure \ref{fig:flow_boxplot} shows the variations in link flows over the three networks under high disruption. The average coefficients of variation of actual link flow for three networks are 0.297, 0.242, and 0.201, respectively. Furthermore, Figure \ref{fig:urban_flow_loss} shows the predicted value and ground truth of the link flows on multiple samples in the Anaheim network under the LMH-LMH setting. In total, the flow-capacity ratio on 10,000 edges is predicted using HetGAT, GAT, and GCN, respectively. Also, to compare the predictions with ground truth, ccorrelation coefficients and pairwise comparisons are shown in Figure~\ref{fig:urban_flow_loss}, indicating that HetGAT outperforms  GAT and GCN. 

Table \ref{tab:urban_mae} summarizes the prediction performance of all methods under different settings, and shows that HetGAT, compared to other models, offers better performance. When the graph size increases, the proposed model maintains a relatively low MAE compared to GCN and GAT. For instance, in the EMA network, HetGAT offers flow MAEs that are 39.5\%, 56.4\%, and 38.8\% lower than the second best result, in LMH-LMH, L-M, and M-H settings, respectively. In the Anaheim network, HetGAT offers flow MAEs that are 27.2\%, 47.4\%, and 33.1\% lower than the second-best result in LMH-LMH, L-M, and M-H settings, respectively.  This shows that the inclusion of virtual links can assist GNN models in better learning the traffic flow patterns. In addition to the prediction accuracy, the training time is also an important factor in evaluating the efficiency and practicality of machine learning models. The computational time is mainly threefold: time for solving UE-TAP, GNN training time, and GNN inference time. For Sioux Falls, EMA, and Anaheim networks, the time for solving UE-TAP is 56.8, 575.2, and 2612.5 min, respectively. In comparison, the training time of HetGAT is 26.8, 28.9, and 59.7 min, respectively. Moreover, for every 1000 graphs, the inference time of the proposed model is notably efficient at 0.13, 0.15, and 0.31 min, respectively. The computational time of each component demonstrates the computational efficiency of the proposed model for solving traffic assignment problems.

\begin{table}[hbt!]
\centering
\caption{Performance comparison of HetGAT with benchmark methods. Three different settings are included: LMH-LMH, L-M, M-H. The mean absolute error, root mean square error, and normalized conservation loss are used to evaluate the prediction performance on the testing set.}
\label{tab:urban_mae}
\renewcommand{\arraystretch}{1.15}
\resizebox{\textwidth}{!}{%
\begin{tabular}{ccccccc|ccccc|ccccc}
\hline
\multirow{3}{*}{Network} & \multirow{3}{*}{Model} & \multicolumn{5}{c}{LMH-LMH} & \multicolumn{5}{c}{L-M} & \multicolumn{5}{c}{M-H} \\ \cline{3-17} 
 &  & \multicolumn{2}{c}{\begin{tabular}[c]{@{}c@{}}Flow\\ ($1 \times 10^2$)\end{tabular}} & \multicolumn{2}{c}{\begin{tabular}[c]{@{}c@{}}Link\\ utilization\end{tabular}} & \multirow{2}{*}{$\tilde{L}_c$} & \multicolumn{2}{c}{\begin{tabular}[c]{@{}c@{}}Flow\\ ($1 \times 10^2$)\end{tabular}} & \multicolumn{2}{c}{\begin{tabular}[c]{@{}c@{}}Link\\ utilization\end{tabular}} & \multirow{2}{*}{$\tilde{L}_c$} & \multicolumn{2}{c}{\begin{tabular}[c]{@{}c@{}}Flow\\ ($1 \times 10^2$)\end{tabular}} & \multicolumn{2}{c}{\begin{tabular}[c]{@{}c@{}}Link\\ utilization\end{tabular}} & \multirow{2}{*}{$\tilde{L}_c$} \\ \cline{3-6} \cline{8-11} \cline{13-16}
 &  & MAE & RMSE & MAE & RMSE &  & MAE & RMSE & MAE & RMSE &  & MAE & RMSE & MAE & RMSE &  \\ \hline
\multirow{4}{*}{Sioux Falls} & FCNN & 19.31 & 30.22 & 16.29\% & 26.39\% & 0.38 & 18.64 & 27.31 & 15.87\% & 24.55\% & 0.37 & 32.85 & 48.08 & 30.09\% & 44.99\% & 0.70 \\
 & GCN & 5.62 & 8.80 & 4.78\% & 7.74\% & 0.11 & 5.43 & 7.95 & 4.62\% & 7.15\% & 0.14 & 9.57 & 14.00 & 8.64\% & 12.87\% & 0.22 \\
 & GAT & 5.60 & 8.75 & 4.74\% & 7.68\% & 0.11 & 5.54 & 8.18 & 4.62\% & 7.15\% & 0.11 & 9.55 & 13.99 & 8.76\% & 13.10\% & 0.20 \\
 & HetGAT & \textbf{3.58} & \textbf{5.56} & \textbf{2.99\%} & \textbf{4.74\%} & \textbf{0.07} & \textbf{3.73} & \textbf{5.39} & \textbf{2.99\%} & \textbf{4.36\%} & \textbf{0.07} & \textbf{7.01} & \textbf{10.39} & \textbf{6.32\%} & \textbf{9.52\%} & \textbf{0.08} \\ \hline
\multirow{4}{*}{EMA} & FCNN & 7.51 & 14.82 & 25.56\% & 63.89\% & 0.81 & 9.04 & 17.70 & 32.36\% & 61.87\% & 0.46 & 13.82 & 24.68 & 60.47\% & 122.94\% & 0.62 \\
 & GCN & 1.99 & 3.92 & 7.62\% & 18.32\% & 0.17 & 2.39 & 4.69 & 7.78\% & 14.65\% & 0.08 & 3.66 & 6.53 & 14.35\% & 30.33\% & 0.12 \\
 & GAT & 1.77 & 3.62 & 6.77\% & 16.91\% & 0.21 & 2.39 & 4.69 & 8.57\% & 16.38\% & 0.12 & 3.83 & 6.87 & 16.01\% & 32.54\% & 0.17 \\
 & HetGAT & \textbf{0.98} & \textbf{2.21} & \textbf{3.48\%} & \textbf{8.08\%} & \textbf{0.07} & \textbf{1.05} & \textbf{2.06} & \textbf{3.52\%} & \textbf{6.85\%} & \textbf{0.03} & \textbf{2.49} & \textbf{4.79} & \textbf{9.24\%} & \textbf{20.50\%} & \textbf{0.07} \\ \hline
\multirow{4}{*}{Anaheim} & FCNN & 10.61 & 16.73 & 24.72\% & 45.03\% & 0.81 & 15.82 & 28.59 & 38.44\% & 69.66\% & 0.23 & 18.74 & 29.82 & 49.70\% & 84.45\% & 0.25 \\
 & GCN & 2.18 & 3.44 & 6.54\% & 11.53\% & 0.29 & 3.25 & 5.88 & 9.31\% & 17.09\% & 0.07 & 3.85 & 6.13 & 12.37\% & 20.74\% & 0.08 \\
 & GAT & 1.47 & 2.55 & 4.77\% & 8.69\% & 0.16 & 2.32 & 4.18 & 7.42\% & 13.45\% & 0.05 & 3.31 & 5.14 & 9.59\% & 16.30\% & 0.05 \\
 & HetGAT & \textbf{1.04} & \textbf{1.81} & \textbf{2.97\%} & \textbf{5.46\%} & \textbf{0.08} & \textbf{1.19} & \textbf{1.85} & \textbf{3.31\%} & \textbf{5.45\%} & \textbf{0.02} & \textbf{2.17} & \textbf{3.37} & \textbf{6.66\%} & \textbf{10.85\%} & \textbf{0.03} \\ \hline
\end{tabular}%
}
\end{table}

\subsubsection{Incomplete OD demand data}
As an additional experiment, we consider a realistic scenario where the regional OD demand values are incomplete by introducing a random binary mask to the original OD demand, More specifically, given a specific missing ratio, we randomly select a number of OD pairs and mask their corresponding OD demand values as zeros in the input node feature. In this way, the model is expected to learn the inherent patterns and structures of the transportation network, even when some of the demand information is missing. The effectiveness of the proposed model under incomplete OD demand scenarios will be evaluated by comparing the predicted traffic flows against the ground truth data obtained from the complete OD demand. Three missing ratios are considered in the experiment: 20\%, 30\%, and 40\%. The training setting and the hyperparameters remain the same as those in the aforementioned experiments. Table \ref{tab:incomplete_mae} summarizes the results of prediction performance under different missing rate scenarios for the LMH-LMH setting. The FCNN model is not considered in these experiments because of their very poor performance in the previous experiments under full OD demand. HetGAT still outperforms GAT and GCN under different networks and different missing ratios. additionally, the flow predictions by HetGAT have relatively better compliance with the flow conservation law compared with GAT and GCN. The superior performance of HetGAT in scenarios with missing OD demand can be attributed to inclusion of virtual OD links and an adaptive attention mechanism, which captures the inherent correlation between OD demand features and link flows from a lower-dimensional embedding space. With the help of the V-encoder and R-encoder, HetGAT seeks to align incomplete demand embeddings with complete OD demand embeddings in the embedding space. This alignment enables the model to effectively reconstruct full OD demands from partial data, thereby enhancing the accuracy of its predictions.

\begin{table}[htb!]
\centering
\caption{Comparison of the performance of HetGAT with benchmark under incomplete OD demand. Three missing ratios are considered in the experiments: 20\%, 30\%, and 40\%. The mean absolute error, root mean square error, and normalized conservation loss, are used to evaluate the prediction performance on the testing set.}
\label{tab:incomplete_mae}
\renewcommand{\arraystretch}{1.15}
\resizebox{\textwidth}{!}{%
\begin{tabular}{ccccccc|ccccc|ccccc}
\hline
\multirow{3}{*}{Network} & \multirow{3}{*}{Model} & \multicolumn{5}{c}{Missing ratio = 20\%} & \multicolumn{5}{c}{Missing ratio = 30\%} & \multicolumn{5}{c}{Missing ratio =40\%} \\ \cline{3-17} 
 &  & \multicolumn{2}{c}{\begin{tabular}[c]{@{}c@{}}Flow\\ ($1 \times 10^2$)\end{tabular}} & \multicolumn{2}{c}{\begin{tabular}[c]{@{}c@{}}Link\\ utilization\end{tabular}} & \multirow{2}{*}{$\tilde{L}_c$} & \multicolumn{2}{c}{\begin{tabular}[c]{@{}c@{}}Flow\\ ($1 \times 10^2$)\end{tabular}} & \multicolumn{2}{c}{\begin{tabular}[c]{@{}c@{}}Link\\ utilization\end{tabular}} & \multirow{2}{*}{$\tilde{L}_c$} & \multicolumn{2}{c}{\begin{tabular}[c]{@{}c@{}}Flow\\ ($1 \times 10^2$)\end{tabular}} & \multicolumn{2}{c}{\begin{tabular}[c]{@{}c@{}}Link\\ utilization\end{tabular}} & \multirow{2}{*}{$\tilde{L}_c$} \\ \cline{3-6} \cline{8-11} \cline{13-16}
 &  & MAE & RMSE & MAE & RMSE &  & MAE & RMSE & MAE & RMSE &  & MAE & RMSE & MAE & RMSE &  \\ \hline
\multirow{3}{*}{Sioux Falls} & GCN & 6.50 & 10.00 & 5.39\% & 8.62\% & 0.15 & 6.54 & 10.02 & 5.72\% & 9.19\% & 0.15 & 7.54 & 11.06 & 6.68\% & 10.60\% & 0.16 \\
 & GAT & 6.64 & 10.11 & 5.83\% & 9.25\% & 0.14 & 7.19 & 10.50 & 6.11\% & 9.45\% & 0.14 & 6.68 & 10.13 & 5.78\% & 9.20\% & 0.15 \\
 & HetGAT & \textbf{4.05} & \textbf{6.04} & \textbf{3.33\%} & \textbf{5.06\%} & \textbf{0.08} & \textbf{4.13} & \textbf{6.14} & \textbf{3.38\%} & \textbf{5.13\%} & \textbf{0.09} & \textbf{4.15} & \textbf{6.08} & \textbf{3.45\%} & \textbf{5.16\%} & \textbf{0.09} \\ \hline
\multirow{3}{*}{EMA} & GCN & 2.14 & 4.11 & 8.06\% & 16.94\% & 0.24 & 2.15 & 4.10 & 7.87\% & 16.47\% & 0.25 & 2.14 & 4.12 & 8.03\% & 17.09\% & 0.26 \\
 & GAT & 1.87 & 3.71 & 6.89\% & 16.65\% & 0.18 & 1.86 & 3.73 & 7.00\% & 16.77\% & 0.19 & 1.87 & 3.73 & 7.07\% & 16.97\% & 0.19 \\
 & HetGAT & \textbf{1.15} & \textbf{2.37} & \textbf{3.98\%} & \textbf{8.40\%} & \textbf{0.08} & \textbf{1.15} & \textbf{2.37} & \textbf{4.00\%} & \textbf{8.24\%} & \textbf{0.08} & \textbf{1.21} & \textbf{2.41} & \textbf{4.17\%} & \textbf{8.67\%} & \textbf{0.09} \\ \hline
\multirow{3}{*}{Anaheim} & GCN & 2.16 & 3.38 & 6.31\% & 10.86\% & 0.14 & 2.24 & 3.43 & 6.65\% & 11.18\% & 0.14 & 2.28 & 3.56 & 6.82\% & 11.72\% & 0.15 \\
 & GAT & 1.62 & 2.70 & 4.80\% & 8.70\% & 0.12 & 1.74 & 2.85 & 5.21\% & 9.34\% & 0.13 & 1.68 & 2.78 & 4.98\% & 8.97\% & 0.12 \\
 & HetGAT & \textbf{1.11} & \textbf{1.91} & \textbf{3.17\%} & \textbf{5.78\%} & \textbf{0.06} & \textbf{1.08} & \textbf{1.87} & \textbf{3.08\%} & \textbf{5.65\%} & \textbf{0.05} & \textbf{1.09} & \textbf{1.89} & \textbf{3.12\%} & \textbf{5.72\%} & \textbf{0.05} \\ \hline
\end{tabular}%
}
\end{table}

\subsection{Experiments on Generalized Synthetic Networks}
\label{sec:synthetic}
In this section, unlike the experiments in Section \ref{sec:urban}, which involved training and testing on an identical network topology,  we examine the generalization capability of the proposed model to networks with varied topologies. In particular, we consider real-world scenarios in which certain links in the network are fully closed due to governmental directives or catastrophic events such as bridge collapses, leading to significant alterations in network topology. Another scenario for topology alternation is when cities consider network expansion to better serve increased mobility demands due to current urbanization trends. Under all these scenarios, the resulting urban networks may exhibit both commonalities and disparities in their topologies \citep{rodrigue2020geography}. This is while training models separately for each distinct network requires substantial time and effort. Motivated by the aforementioned considerations, our aim is to explore the generalization ability of our  HetGAT model over varying topologies. In this work, we conduct experiments on two sets of networks: (1) modified urban networks of Section \ref{sec:urban}; (2) synthetic networks.  The first set, includes networks that are modifications of the original Sioux Falls, EMA, and Anaheim networks by adding and removing links.

\subsubsection{Modified urban networks}
 In this section, we generate synthetic networks by modifying the three urban networks used in previous seciton. For each of these networks, we generate 20 unique topological variations.  Additionally, we incorporate three configurations of OD demand in our experiments: complete OD setting, 20\% and 40\% incomplete OD setting. The training setting and the hyperparameters remain the same as those in the aforementioned experiments in Section \ref{sec:urban}. The results are presented in Table \ref{tab:multigraph}. It can be seen that compared to the results in Table \ref{tab:urban_mae}, HetGAT exhibits comparable levels of effectiveness for all networks. Additionally, in comparison to other baseline models, our proposed HetGAT model consistently outperforms both GAT and GCN over all networks, under both complete and incomplete OD scenarios.  It is noteworthy that compared to the homogeneous GNN model, since our model has captured the influence of OD pairs via virtual links, the HetGAT model is more expressive and can better capture and learn the impact of topology alterations on flow distributions.

\begin{table}[htb!]
\centering
\caption{Performance comparison of HetGAT with benchmark methods on modified urban transportation network with link addition and removal. The mean absolute error, root mean square error, and normalized conservation loss are used to evaluate the prediction
performance on the testing set.}
\label{tab:multigraph}
\renewcommand{\arraystretch}{1.15}
\resizebox{\textwidth}{!}{%
\begin{tabular}{ccccccc|ccccc|ccccc}
\hline
\multirow{3}{*}{Network} & \multirow{3}{*}{Model} & \multicolumn{5}{c}{Missing Ratio = 0\%} & \multicolumn{5}{c}{Missing Ratio = 20\%} & \multicolumn{5}{c}{Missing Ratio = 40\%} \\ \cline{3-17} 
 &  & \multicolumn{2}{c}{\begin{tabular}[c]{@{}c@{}}Flow\\ ($1 \times 10^2$)\end{tabular}} & \multicolumn{2}{c}{\begin{tabular}[c]{@{}c@{}}Link\\ utilization\end{tabular}} & \multirow{2}{*}{$\tilde{L}_c$} & \multicolumn{2}{c}{\begin{tabular}[c]{@{}c@{}}Flow\\ ($1 \times 10^2$)\end{tabular}} & \multicolumn{2}{c}{\begin{tabular}[c]{@{}c@{}}Link\\ utilization\end{tabular}} & \multirow{2}{*}{$\tilde{L}_c$} & \multicolumn{2}{c}{\begin{tabular}[c]{@{}c@{}}Flow\\ ($1 \times 10^2$)\end{tabular}} & \multicolumn{2}{c}{\begin{tabular}[c]{@{}c@{}}Link\\ utilization\end{tabular}} & \multirow{2}{*}{$\tilde{L}_c$} \\ \cline{3-6} \cline{8-11} \cline{13-16}
 &  & MAE & RMSE & MAE & RMSE &  & MAE & RMSE & MAE & RMSE &  & MAE & RMSE & MAE & RMSE &  \\ \hline
\multirow{3}{*}{Sioux} & GCN & 10.12 & 13.97 & 8.31\% & 11.77\% & 0.15 & 10.96 & 14.82 & 9.22\% & 13.05\% & 0.15 & 10.94 & 15.04 & 9.05\% & 12.86\% & 0.15 \\
 & GAT & 7.16 & 9.89 & 6.10\% & 8.77\% & 0.13 & 10.78 & 14.40 & 9.11\% & 12.77\% & 0.14 & 12.20 & 16.42 & 10.34\% & 14.58\% & 0.17 \\
 & HetGAT & \textbf{3.04} & \textbf{4.22} & \textbf{2.56\%} & \textbf{3.75\%} & \textbf{0.07} & \textbf{3.33} & \textbf{4.53} & \textbf{2.79\%} & \textbf{3.96\%} & \textbf{0.07} & \textbf{3.46} & \textbf{4.69} & \textbf{2.88\%} & \textbf{4.08\%} & \textbf{0.08} \\ \hline
\multirow{3}{*}{EMA} & GCN & 4.60 & 7.82 & 15.11\% & 26.42\% & 0.26 & 4.09 & 6.68 & 13.97\% & 24.71\% & 0.24 & 4.04 & 6.57 & 15.24\% & 31.99\% & 0.23 \\
 & GAT & 2.70 & 4.50 & 9.97\% & 18.21\% & 0.14 & 2.73 & 4.55 & 10.04\% & 20.05\% & 0.15 & 2.86 & 4.72 & 10.50\% & 19.94\% & 0.15 \\
 & HetGAT & \textbf{1.28} & \textbf{2.39} & \textbf{4.09\%} & \textbf{7.91\%} & \textbf{0.06} & \textbf{1.34} & \textbf{2.47} & \textbf{4.29\%} & \textbf{8.17\%} & \textbf{0.07} & \textbf{1.42} & \textbf{2.57} & \textbf{4.56\%} & \textbf{8.42\%} & \textbf{0.07} \\ \hline
\multirow{3}{*}{Anaheim} & GCN & 6.02 & 9.93 & 16.29\% & 26.78\% & 0.40 & 5.93 & 9.83 & 15.98\% & 26.40\% & 0.31 & 6.04 & 9.99 & 16.24\% & 26.73\% & 0.37 \\
 & GAT & 5.43 & 9.16 & 14.94\% & 25.20\% & 0.25 & 5.37 & 9.02 & 14.83\% & 25.01\% & 0.40 & 5.53 & 9.28 & 15.23\% & 25.60\% & 0.34 \\
 & HetGAT & \textbf{1.14} & \textbf{2.22} & \textbf{3.29\%} & \textbf{6.79\%} & \textbf{0.09} & \textbf{1.15} & \textbf{2.08} & \textbf{3.34\%} & \textbf{6.49\%} & \textbf{0.09} & \textbf{1.29} & \textbf{2.38} & \textbf{3.71\%} & \textbf{7.40\%} & \textbf{0.10} \\ \hline
\end{tabular}%
}
\end{table}

\subsubsection{Randomly generated graphs and generalization to variable graph size}
\begin{figure}[h]
\centering
\begin{subfigure}[b]{0.33\textwidth}
    \centering
    \includegraphics[width=\textwidth]{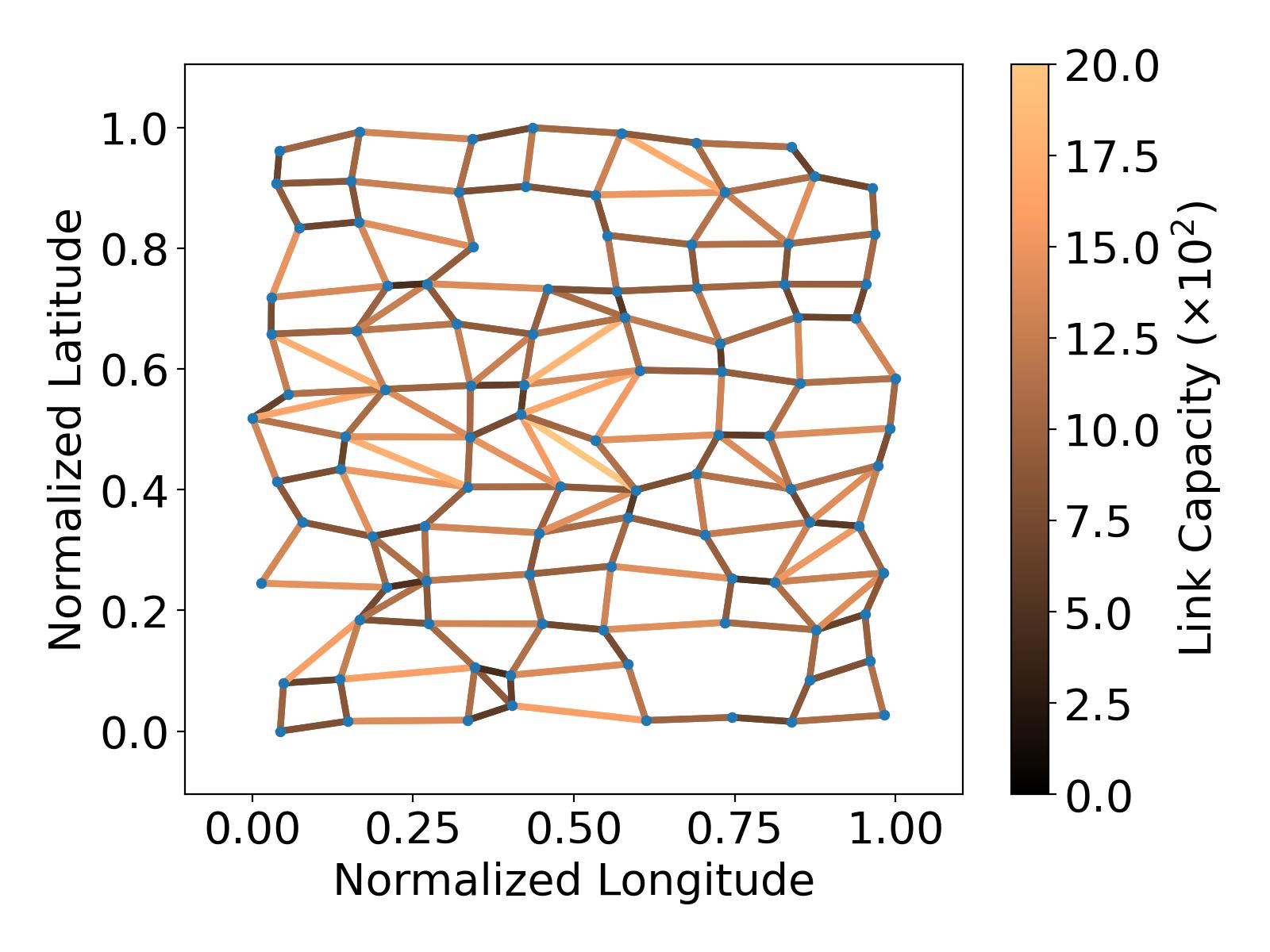}
    \label{fig:sample_1}
\end{subfigure}
\hfill
\begin{subfigure}[b]{0.32\textwidth}
    \centering
    \includegraphics[width=\textwidth]{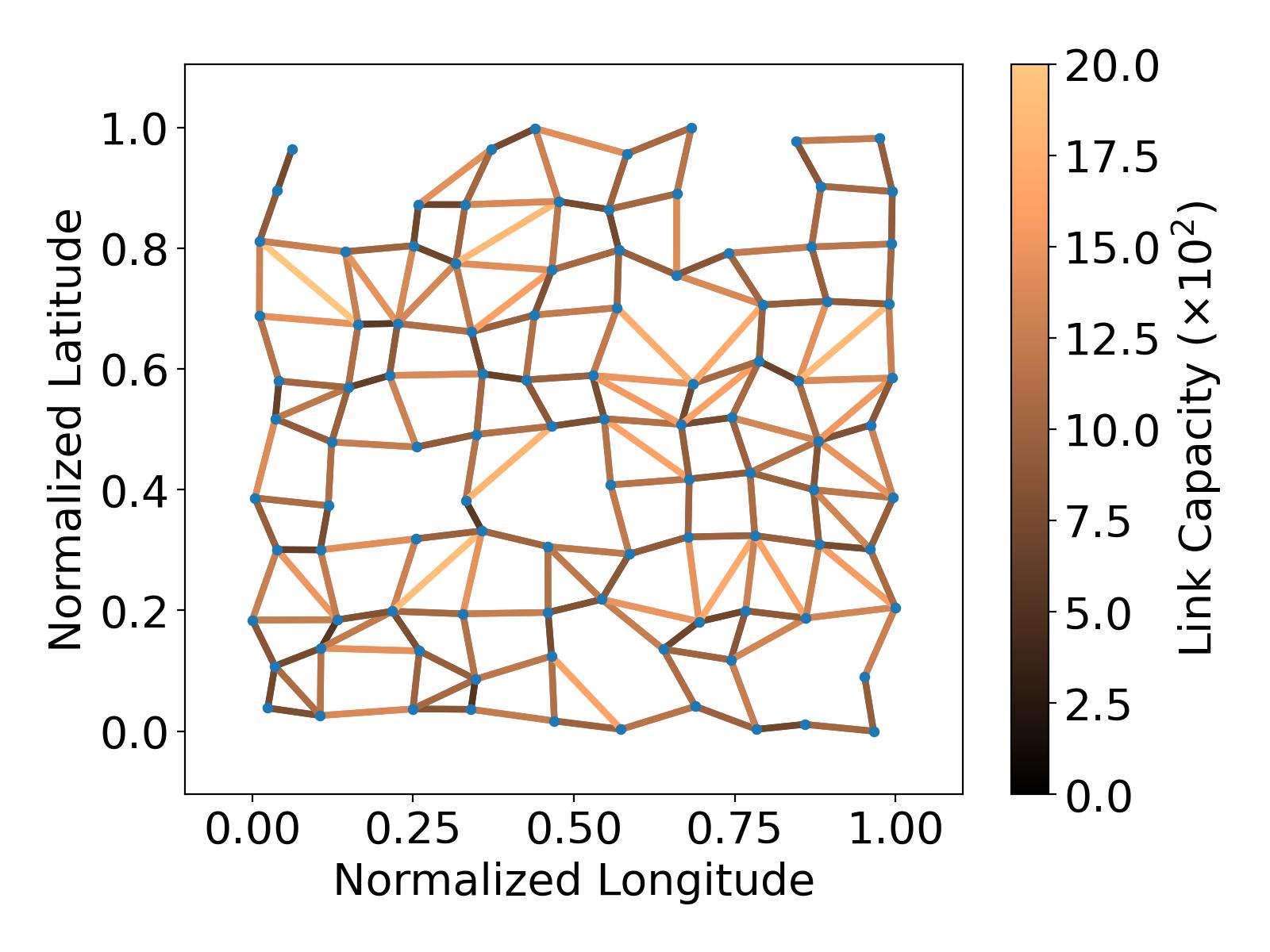}
    \label{fig:sample_2}
\end{subfigure}
\hfill
\begin{subfigure}[b]{0.33\textwidth}
    \centering
    \includegraphics[width=\textwidth]{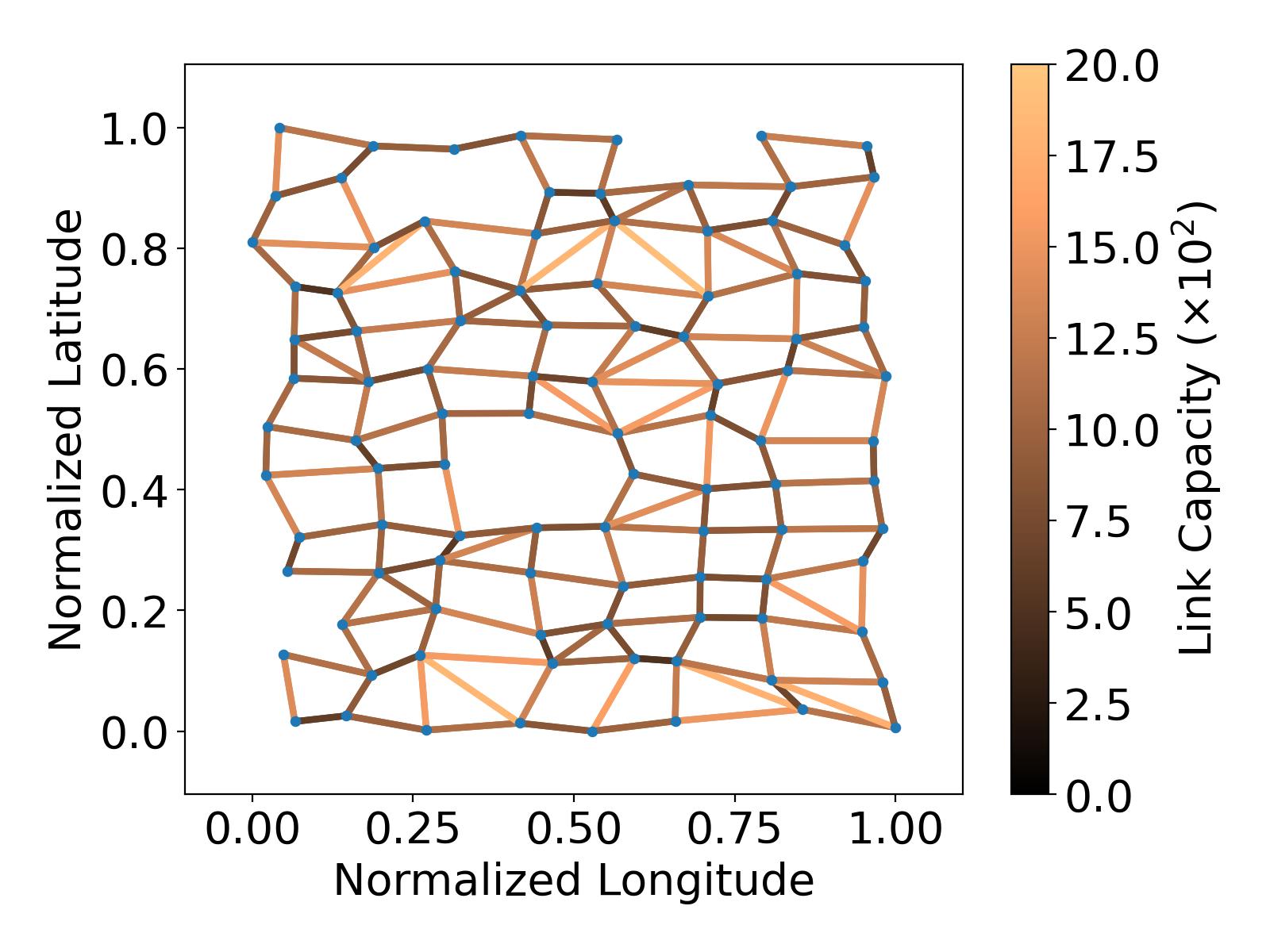}
    \label{fig:sample_3}
\end{subfigure}
    \caption{The illustrations of sampled generalized synthetic networks with the network size of 100. The link color represents the link capacity of each link.}
    \label{fig:synthetic_network}
\end{figure}

\begin{figure}[!htb]
\centering
\begin{subfigure}[b]{0.3\textwidth}
    \centering
    \includegraphics[width=\textwidth]{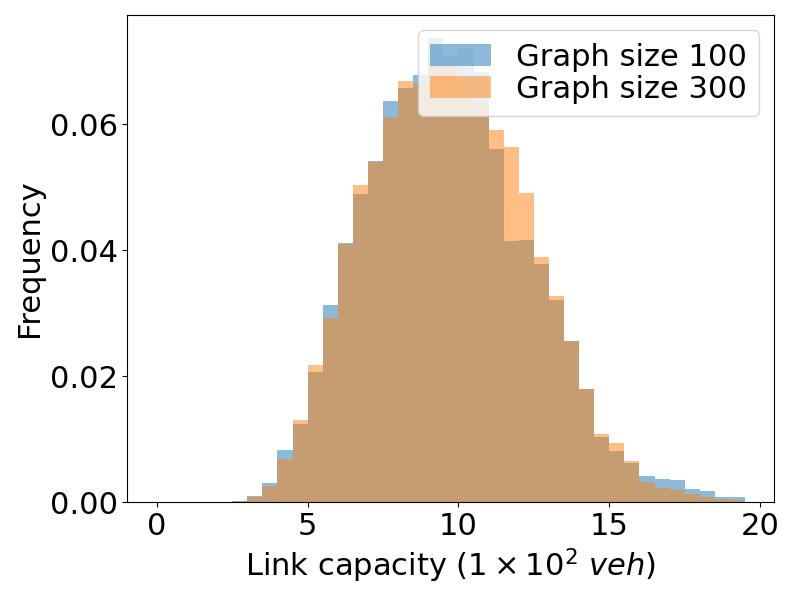}
    \label{fig:all_random_link_capacity_variation}
\end{subfigure}
\quad
\begin{subfigure}[b]{0.3\textwidth}
    \centering
    \includegraphics[width=\textwidth]{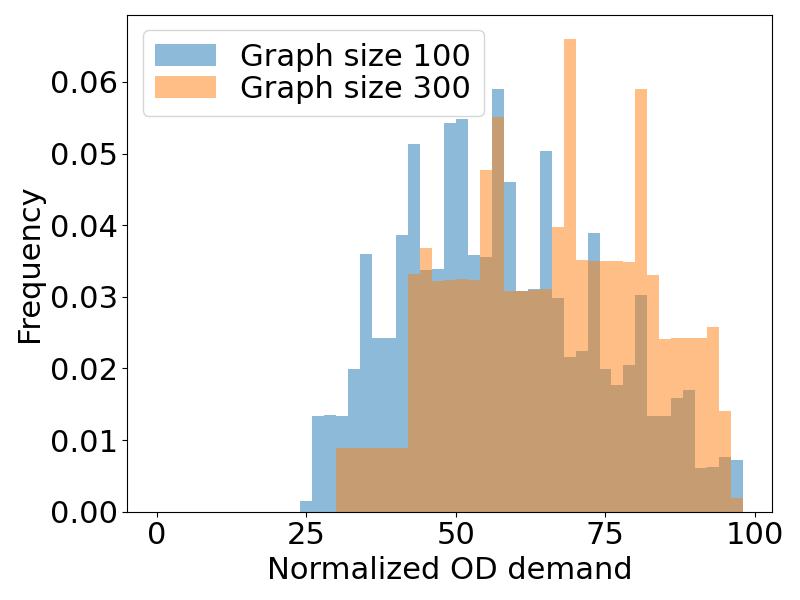}
    \label{fig:all_random_od_variation}
\end{subfigure}
\caption{The histogram of the link capacity and normalized OD demand in the training and testing data of synthetic networks. Two sizes of datasets are included: 100 and 300.}
\label{fig:variation_histogram_random}
\end{figure}

The second set of synthetic networks are randomly generated   by starting with a grid graph and then adding links between randomly selected nodes. Furthermore, to emulate the real road network, a number of nodes and edges are randomly removed until the number of nodes in the graph reaches a predefined threshold. Two sets of synthetic networks are considered: one with 100 nodes and another with 300 nodes. For each graph size, 20 different graph topologies are generated. Three of these randomly generated graphs of size 100 are shown in Figure \ref{fig:synthetic_network}. The OD demand and the link capacity are also randomly generated using the scaling factor according to Equations \ref{eq:od_demand} and \ref{eq:ca_demand}, respectively. To demonstrate the diversity of these examples, similar to Section \ref{sec:urban}, the histograms of the link capacities and the OD demands in the generated dataset are shown in Figure \ref{fig:variation_histogram_random}, indicating the training and testing data are sufficiently diverse to cover different scenarios.

In this section, we investigate the generalization of the proposed method when graph sizes in the training and test cases are different. In particular, we consider two different graph sizes, with the node numbers $N=100$ and $N=300$.  It should be noted that a node feature in our proposed HetGAT model has a dimension  equal to the total number of nodes in the graph. Therefore, to handle variable graph sizes in the dataset, we need to make an adjustment to the model. To this end, we propose two training strategies for  HetGAT:

\begin{itemize}
    \item Transfer learning: In this approach, we use the model trained from previous experiments as a pre-trained model and let the parameters of its preprocessing and final layers be re-trained. The remaining parameters will be frozen and unchanged during this re-training. So, for subsequent training involving graphs of the new size, the optimization will be relatively fast as it is only done for the parameters in two layers, rather than the entire model. Two cases are studied: when testing is done on graphs with $N=100$ and training is done with $N=300$, and vice versa. 
    \item Homogenized training: In this approach, we set a ceiling (maximum) for the node number  $N_{\max}$, anticipating the largest graph that can ever be handled. Then, for all the graphs containing fewer nodes, we add enough dummy nodes to the graph for it size to reach $N_{\max}$. This approach ensures uniform formulation and architecture across different graphs, allowing a single model to be used for graphs of varying sizes. To numerically study this strategy, the model is trained once  using graphs with sizes $N=100$ and $N=300$. This trained model is then tested on graphs with $N=100$ and graphs with $N=300$ separately. 
\end{itemize}

In order to evaluate the effectiveness of our proposed model compared to baseline models, we compare our results in a standard training, where training and test graphs are of the same size. That means for test graphs with $N=100$, HetGAT was trained in the standard way of the previous section, with only graphs with $N=100$. Similarly, training and test graphs of size $N=300$ were used for the second set of ``standard training'' results. 

The training setting and the hyperparameters remain the same as those in the previous experiments. The prediction performance metrics for the testing sets are presented in Table \ref{tab:generalized_mae_100} and \ref{tab:generalized_mae_300}. First,  it can be seen that  the proposed HetGAT model achieves the best accuracy levels compared to other model architectures on randomly generated graphs when same graph size is used in training and test. Furthermore, for the cases with variable graph sizes, even though the performance decline under the transfer and homogenized learning setting, the proposed HetGAT model maintains a competitive edge, outperforming baseline models such as GAT and GCN.

\begin{table}[htb!]
\centering
\caption{Comparison of the performance of HetGAT with that of GAT and GCN on generalized synthetic networks. The graph size in testing dataset is 100. We consider three different training strategies: standard training, transfer learning, and homogenized learning.}
\label{tab:generalized_mae_100}
\renewcommand{\arraystretch}{1.15}
\resizebox{\textwidth}{!}{%
\begin{tabular}{ccccccccccccccccc}
\hline
\multirow{4}{*}{\begin{tabular}[c]{@{}c@{}}Testing\\ dataset\end{tabular}} & \multirow{4}{*}{Model} & \multicolumn{5}{c}{Standard training} & \multicolumn{5}{c}{Transfer learning} & \multicolumn{5}{c}{Homogenized learning} \\ \cline{3-17} 
&  & \multicolumn{5}{c}{Training N = 100} & \multicolumn{5}{c}{Training N = 300} & \multicolumn{5}{c}{Training N = \{100, 300\}} \\ \cline{3-17} 
&  & \multicolumn{2}{c}{\begin{tabular}[c]{@{}c@{}}Flow\\ ($1 \times 10^2$)\end{tabular}} & \multicolumn{2}{c}{\begin{tabular}[c]{@{}c@{}}Link\\ utilization\end{tabular}} & \multirow{2}{*}{$\tilde{L}_c$} & \multicolumn{2}{c}{\begin{tabular}[c]{@{}c@{}}Flow\\ ($1 \times 10^2$)\end{tabular}} & \multicolumn{2}{c}{\begin{tabular}[c]{@{}c@{}}Link\\ utilization\end{tabular}} & \multirow{2}{*}{$\tilde{L}_c$} & \multicolumn{2}{c}{\begin{tabular}[c]{@{}c@{}}Flow\\ ($1 \times 10^2$)\end{tabular}} & \multicolumn{2}{c}{\begin{tabular}[c]{@{}c@{}}Link\\ utilization\end{tabular}} & \multirow{2}{*}{$\tilde{L}_c$} \\ \cline{3-6} \cline{8-11} \cline{13-16}
 &  & MAE & RMSE & MAE & RMSE &  & MAE & RMSE & MAE & RMSE &  & MAE & RMSE & MAE & RMSE &  \\ \hline
\multirow{3}{*}{N=100} & GCN & 1.62 & 2.40 & 17.85\% & 26.89\% & 0.38 & 1.82 & 2.63 & 20.09\% & 29.70\% & 0.10 & 1.45 & 2.08 & 15.76\% & 23.11\% & 0.11 \\
 & GAT & 1.68 & 2.48 & 18.40\% & 27.49\% & 0.33 & 1.92 & 2.79 & 21.11\% & 31.30\% & 0.10 & 1.63 & 2.42 & 17.90\% & 27.59\% & 0.11 \\
 & HetGAT & \textbf{0.25} & \textbf{0.38} & \textbf{2.75\%} & \textbf{4.27\%} & \textbf{0.19} & \textbf{0.38} & \textbf{0.56} & \textbf{4.18\%} & \textbf{6.29\%} & \textbf{0.09} & \textbf{0.74} & \textbf{1.16} & \textbf{8.46\%} & \textbf{14.35\%} & \textbf{0.09} \\ \hline
\end{tabular}%
}
\end{table}

\begin{table}[htb!]
\centering
\caption{Comparison of the performance of HetGAT with that of GAT and GCN on generalized synthetic networks. The graph size in testing dataset is 300. We consider three different training strategies: standard training, transfer learning, and homogenized learning.}
\label{tab:generalized_mae_300}
\renewcommand{\arraystretch}{1.15}
\resizebox{\textwidth}{!}{%
\begin{tabular}{ccccccccccccccccc}
\hline
\multirow{4}{*}{\begin{tabular}[c]{@{}c@{}}Testing\\ dataset\end{tabular}} & \multirow{4}{*}{Model} & \multicolumn{5}{c}{Standard training} & \multicolumn{5}{c}{Transfer learning} & \multicolumn{5}{c}{Homogenized learning} \\ \cline{3-17} 
&  & \multicolumn{5}{c}{Training N = 300} & \multicolumn{5}{c}{Training N = 100} & \multicolumn{5}{c}{Training N = \{100, 300\}} \\ \cline{3-17} 
&  & \multicolumn{2}{c}{\begin{tabular}[c]{@{}c@{}}Flow\\ ($1 \times 10^2$)\end{tabular}} & \multicolumn{2}{c}{\begin{tabular}[c]{@{}c@{}}Link\\ utilization\end{tabular}} & \multirow{2}{*}{$\tilde{L}_c$} & \multicolumn{2}{c}{\begin{tabular}[c]{@{}c@{}}Flow\\ ($1 \times 10^2$)\end{tabular}} & \multicolumn{2}{c}{\begin{tabular}[c]{@{}c@{}}Link\\ utilization\end{tabular}} & \multirow{2}{*}{$\tilde{L}_c$} & \multicolumn{2}{c}{\begin{tabular}[c]{@{}c@{}}Flow\\ ($1 \times 10^2$)\end{tabular}} & \multicolumn{2}{c}{\begin{tabular}[c]{@{}c@{}}Link\\ utilization\end{tabular}} & \multirow{2}{*}{$\tilde{L}_c$} \\ \cline{3-6} \cline{8-11} \cline{13-16}
 &  & MAE & RMSE & MAE & RMSE &  & MAE & RMSE & MAE & RMSE &  & MAE & RMSE & MAE & RMSE &  \\ \hline
\multirow{3}{*}{N=300} & GCN & 2.97 & 4.11 & 31.87\% & 44.14\% & 0.15 & 3.09 & 4.28 & 33.17\% & 45.97\% & 0.13 & 3.04 & 4.23 & 32.65\% & 45.32\% & 0.14 \\
 & GAT & 2.81 & 3.89 & 30.10\% & 41.58\% & 0.18 & 2.94 & 4.05 & 31.43\% & 43.33\% & 0.15 & 2.85 & 3.94 & 30.56\% & 42.19\% & 0.16 \\
 & HetGAT & \textbf{0.46} & \textbf{0.69} & \textbf{4.99\%} & \textbf{7.51\%} & \textbf{0.11} & \textbf{0.78} & \textbf{1.17} & \textbf{8.45\%} & \textbf{12.89\%} & \textbf{0.06} & \textbf{0.52} & \textbf{0.77} & \textbf{5.63\%} & \textbf{8.45\%} & \textbf{0.13} \\ \hline
\end{tabular}%
}
\end{table}

\section{Conclusion and Discussion}
\label{sec:conslusion}

In this paper, we proposed a novel approach for traffic assignment using an end-to-end heterogeneous graph neural network. Compared to conventional homogeneous graph neural networks, our proposed architecture includes additional virtual links connecting origin-destination node pairs, to better uncover dependencies between link flows and OD demand. We also proposed a novel adaptive graph attention mechanism to effectively capture the semantic and contextual features through different types of links. The extensive experiments on three real-world urban transportation networks showed that the proposed model outperforms other state-of-the-arth models in terms of convergence rate and prediction accuracy. Notably, by introducing two
different training strategies, the proposed heterogeneous graph neural network model can also be generalized to different network topologies, underscoring its potential in real-world scenarios.

The proposed framework in this work can serve as a surrogate model that can significantly accelerate complex optimization tasks in areas such as resource allocation and infrastructure asset management. The significant reduction in computational time allows planners to analyze networks under a large number of scenarios in a more comprehensive and more realistic decision-making process. The generalization advantage of this model is particularly beneficial for network design, especially considering network expansion options. Moreover, it was demonstrated  the HetGAT model can robustly predict link flows from inaccurate OD demand data. In this current version, the proposed HetGAT model only learns and predicts the static traffic flow patterns. As a potential extension of this work, the proposed framework can be extended to learn the dynamic traffic flow patterns. Furthermore, the current proposed GNN model uses training data collected from conventional solvers of static traffic assignment. In future work, we will explore how these models can be trained on traffic data collected from sensors, such as loop detectors, cameras, or GPS devices.

\section{Acknowledgment}
This work was supported in part by the National Science Foundation under Grant CMMI-1752302.

\bibliographystyle{elsarticle-harv} 
\bibliography{cas-refs}

\begin{thebibliography}{34}
\expandafter\ifx\csname natexlab\endcsname\relax\def\natexlab#1{#1}\fi
\providecommand{\url}[1]{\texttt{#1}}
\providecommand{\href}[2]{#2}
\providecommand{\path}[1]{#1}
\providecommand{\DOIprefix}{doi:}
\providecommand{\ArXivprefix}{arXiv:}
\providecommand{\URLprefix}{URL: }
\providecommand{\Pubmedprefix}{pmid:}
\providecommand{\doi}[1]{\href{http://dx.doi.org/#1}{\path{#1}}}
\providecommand{\Pubmed}[1]{\href{pmid:#1}{\path{#1}}}
\providecommand{\bibinfo}[2]{#2}
\ifx\xfnm\relax \def\xfnm[#1]{\unskip,\space#1}\fi
%Type = Article
\bibitem[{Babazadeh et~al.(2020)Babazadeh, Javani, Gentile and
  Florian}]{babazadeh2020reduced}
\bibinfo{author}{Babazadeh, A.}, \bibinfo{author}{Javani, B.},
  \bibinfo{author}{Gentile, G.}, \bibinfo{author}{Florian, M.},
  \bibinfo{year}{2020}.
\newblock \bibinfo{title}{Reduced gradient algorithm for user equilibrium
  traffic assignment problem}.
\newblock \bibinfo{journal}{Transportmetrica A: Transport Science}
  \bibinfo{volume}{16}, \bibinfo{pages}{1111--1135}.
%Type = Article
\bibitem[{Bar-Gera et~al.(2023)Bar-Gera, Stabler and
  Sall}]{bar2021transportation}
\bibinfo{author}{Bar-Gera, H.}, \bibinfo{author}{Stabler, B.},
  \bibinfo{author}{Sall, E.}, \bibinfo{year}{2023}.
\newblock \bibinfo{title}{Transportation networks for research core team}.
\newblock \bibinfo{journal}{Transportation Network Test Problems. Available
  online: https://github.com/bstabler/TransportationNetworks (accessed on May
  14 2023)} .
%Type = Techreport
\bibitem[{Beckmann et~al.(1956)Beckmann, McGuire and
  Winsten}]{beckmann1956studies}
\bibinfo{author}{Beckmann, M.}, \bibinfo{author}{McGuire, C.B.},
  \bibinfo{author}{Winsten, C.B.}, \bibinfo{year}{1956}.
\newblock \bibinfo{title}{Studies in the Economics of Transportation}.
\newblock \bibinfo{type}{Technical Report}.
%Type = Article
\bibitem[{Bedeian and Mossholder(2000)}]{bedeian2000use}
\bibinfo{author}{Bedeian, A.G.}, \bibinfo{author}{Mossholder, K.W.},
  \bibinfo{year}{2000}.
\newblock \bibinfo{title}{On the use of the coefficient of variation as a
  measure of diversity}.
\newblock \bibinfo{journal}{Organizational Research Methods}
  \bibinfo{volume}{3}, \bibinfo{pages}{285--297}.
%Type = Book
\bibitem[{Campbell et~al.(2010)Campbell, Machin and
  Walters}]{campbell2010medical}
\bibinfo{author}{Campbell, M.J.}, \bibinfo{author}{Machin, D.},
  \bibinfo{author}{Walters, S.J.}, \bibinfo{year}{2010}.
\newblock \bibinfo{title}{Medical statistics: a textbook for the health
  sciences}.
\newblock \bibinfo{publisher}{John Wiley \& Sons}.
%Type = Inproceedings
\bibitem[{Cheng and Lin(2024)}]{cheng2024network}
\bibinfo{author}{Cheng, X.}, \bibinfo{author}{Lin, J.}, \bibinfo{year}{2024}.
\newblock \bibinfo{title}{Network equilibrium modeling for long-haul electric
  trucks}, in: \bibinfo{booktitle}{2024 Forum for Innovative Sustainable
  Transportation Systems (FISTS)}, \bibinfo{organization}{IEEE}. pp.
  \bibinfo{pages}{1--6}.
%Type = Inproceedings
\bibitem[{Deng and Ji(2011)}]{deng2011spatiotemporal}
\bibinfo{author}{Deng, Z.}, \bibinfo{author}{Ji, M.}, \bibinfo{year}{2011}.
\newblock \bibinfo{title}{Spatiotemporal structure of taxi services in
  shanghai: Using exploratory spatial data analysis}, in:
  \bibinfo{booktitle}{2011 19th International Conference on Geoinformatics},
  \bibinfo{organization}{IEEE}. pp. \bibinfo{pages}{1--5}.
%Type = Article
\bibitem[{Fan et~al.(2023)Fan, Tang, Ye, Xiao and Zhang}]{fan2023deep}
\bibinfo{author}{Fan, W.}, \bibinfo{author}{Tang, Z.}, \bibinfo{author}{Ye,
  P.}, \bibinfo{author}{Xiao, F.}, \bibinfo{author}{Zhang, J.},
  \bibinfo{year}{2023}.
\newblock \bibinfo{title}{Deep learning-based dynamic traffic assignment with
  incomplete origin--destination data}.
\newblock \bibinfo{journal}{Transportation Research Record}
  \bibinfo{volume}{2677}, \bibinfo{pages}{1340--1356}.
%Type = Inproceedings
\bibitem[{Fu et~al.(2020)Fu, Zhang, Meng and King}]{fu2020magnn}
\bibinfo{author}{Fu, X.}, \bibinfo{author}{Zhang, J.}, \bibinfo{author}{Meng,
  Z.}, \bibinfo{author}{King, I.}, \bibinfo{year}{2020}.
\newblock \bibinfo{title}{Magnn: Metapath aggregated graph neural network for
  heterogeneous graph embedding}, in: \bibinfo{booktitle}{Proceedings of The
  Web Conference 2020}, pp. \bibinfo{pages}{2331--2341}.
%Type = Article
\bibitem[{Fukushima(1984)}]{fukushima1984modified}
\bibinfo{author}{Fukushima, M.}, \bibinfo{year}{1984}.
\newblock \bibinfo{title}{A modified frank-wolfe algorithm for solving the
  traffic assignment problem}.
\newblock \bibinfo{journal}{Transportation Research Part B: Methodological}
  \bibinfo{volume}{18}, \bibinfo{pages}{169--177}.
%Type = Article
\bibitem[{Hornik et~al.(1989)Hornik, Stinchcombe and
  White}]{hornik1989multilayer}
\bibinfo{author}{Hornik, K.}, \bibinfo{author}{Stinchcombe, M.},
  \bibinfo{author}{White, H.}, \bibinfo{year}{1989}.
\newblock \bibinfo{title}{Multilayer feedforward networks are universal
  approximators}.
\newblock \bibinfo{journal}{Neural networks} \bibinfo{volume}{2},
  \bibinfo{pages}{359--366}.
%Type = Article
\bibitem[{Lee et~al.(2003)Lee, Nie and Chen}]{lee2003conjugate}
\bibinfo{author}{Lee, D.H.}, \bibinfo{author}{Nie, Y.}, \bibinfo{author}{Chen,
  A.}, \bibinfo{year}{2003}.
\newblock \bibinfo{title}{A conjugate gradient projection algorithm for the
  traffic assignment problem}.
\newblock \bibinfo{journal}{Mathematical and computer modelling}
  \bibinfo{volume}{37}, \bibinfo{pages}{863--878}.
%Type = Article
\bibitem[{Liu and Meidani(2022)}]{liu2022graph}
\bibinfo{author}{Liu, T.}, \bibinfo{author}{Meidani, H.}, \bibinfo{year}{2022}.
\newblock \bibinfo{title}{Graph neural network surrogate for seismic
  reliability analysis of highway bridge system}.
\newblock \bibinfo{journal}{arXiv preprint arXiv:2210.06404} .
%Type = Incollection
\bibitem[{Liu and Meidani(2023a)}]{liu2023optimizing}
\bibinfo{author}{Liu, T.}, \bibinfo{author}{Meidani, H.},
  \bibinfo{year}{2023}a.
\newblock \bibinfo{title}{Optimizing seismic retrofit of bridges: Integrating
  efficient graph neural network surrogates and transportation equity}, in:
  \bibinfo{booktitle}{Proceedings of Cyber-Physical Systems and Internet of
  Things Week 2023}, pp. \bibinfo{pages}{367--372}.
%Type = Article
\bibitem[{Liu and Meidani(2023b)}]{liu2023physics}
\bibinfo{author}{Liu, T.}, \bibinfo{author}{Meidani, H.},
  \bibinfo{year}{2023}b.
\newblock \bibinfo{title}{Physics-informed neural networks for system
  identification of structural systems with a multiphysics damping model}.
\newblock \bibinfo{journal}{Journal of Engineering Mechanics}
  \bibinfo{volume}{149}, \bibinfo{pages}{04023079}.
%Type = Article
\bibitem[{Liu and Meidani(2024)}]{liu2024neural}
\bibinfo{author}{Liu, T.}, \bibinfo{author}{Meidani, H.}, \bibinfo{year}{2024}.
\newblock \bibinfo{title}{Neural network surrogate models for aerodynamic
  analysis in truck platoons: Implications on autonomous freight delivery}.
\newblock \bibinfo{journal}{International Journal of Transportation Science and
  Technology} .
%Type = Article
\bibitem[{Madadi and de~Almeida~Correia(2024)}]{madadi2024hybrid}
\bibinfo{author}{Madadi, B.}, \bibinfo{author}{de~Almeida~Correia, G.H.},
  \bibinfo{year}{2024}.
\newblock \bibinfo{title}{A hybrid deep-learning-metaheuristic framework for
  bi-level network design problems}.
\newblock \bibinfo{journal}{Expert Systems with Applications}
  \bibinfo{volume}{243}, \bibinfo{pages}{122814}.
%Type = Article
\bibitem[{Nie et~al.(2004)Nie, Zhang and Lee}]{nie2004models}
\bibinfo{author}{Nie, Y.}, \bibinfo{author}{Zhang, H.}, \bibinfo{author}{Lee,
  D.H.}, \bibinfo{year}{2004}.
\newblock \bibinfo{title}{Models and algorithms for the traffic assignment
  problem with link capacity constraints}.
\newblock \bibinfo{journal}{Transportation Research Part B: Methodological}
  \bibinfo{volume}{38}, \bibinfo{pages}{285--312}.
%Type = Article
\bibitem[{Paszke et~al.(2019)Paszke, Gross, Massa, Lerer, Bradbury, Chanan,
  Killeen, Lin, Gimelshein, Antiga et~al.}]{paszke2019pytorch}
\bibinfo{author}{Paszke, A.}, \bibinfo{author}{Gross, S.},
  \bibinfo{author}{Massa, F.}, \bibinfo{author}{Lerer, A.},
  \bibinfo{author}{Bradbury, J.}, \bibinfo{author}{Chanan, G.},
  \bibinfo{author}{Killeen, T.}, \bibinfo{author}{Lin, Z.},
  \bibinfo{author}{Gimelshein, N.}, \bibinfo{author}{Antiga, L.}, et~al.,
  \bibinfo{year}{2019}.
\newblock \bibinfo{title}{Pytorch: An imperative style, high-performance deep
  learning library}.
\newblock \bibinfo{journal}{Advances in neural information processing systems}
  \bibinfo{volume}{32}.
%Type = Article
\bibitem[{Rahman and Hasan(2023)}]{rahman2023data}
\bibinfo{author}{Rahman, R.}, \bibinfo{author}{Hasan, S.},
  \bibinfo{year}{2023}.
\newblock \bibinfo{title}{Data-driven traffic assignment: A novel approach for
  learning traffic flow patterns using graph convolutional neural network}.
\newblock \bibinfo{journal}{Data Science for Transportation}
  \bibinfo{volume}{5}, \bibinfo{pages}{11}.
%Type = Book
\bibitem[{Rodrigue(2020)}]{rodrigue2020geography}
\bibinfo{author}{Rodrigue, J.P.}, \bibinfo{year}{2020}.
\newblock \bibinfo{title}{The geography of transport systems}.
\newblock \bibinfo{publisher}{Routledge}.
%Type = Inproceedings
\bibitem[{Seliverstov et~al.(2017)Seliverstov, Seliverstov, Malygin, Tarantsev,
  Shatalova, Lukomskaya, Tishchenko and
  Elyashevich}]{seliverstov2017development}
\bibinfo{author}{Seliverstov, Y.A.}, \bibinfo{author}{Seliverstov, S.A.},
  \bibinfo{author}{Malygin, I.G.}, \bibinfo{author}{Tarantsev, A.A.},
  \bibinfo{author}{Shatalova, N.V.}, \bibinfo{author}{Lukomskaya, O.Y.},
  \bibinfo{author}{Tishchenko, I.P.}, \bibinfo{author}{Elyashevich, A.M.},
  \bibinfo{year}{2017}.
\newblock \bibinfo{title}{Development of management principles of urban traffic
  under conditions of information uncertainty}, in:
  \bibinfo{booktitle}{Creativity in Intelligent Technologies and Data Science:
  Second Conference, CIT\&DS 2017, Volgograd, Russia, September 12-14, 2017,
  Proceedings 2}, \bibinfo{organization}{Springer}. pp.
  \bibinfo{pages}{399--418}.
%Type = Article
\bibitem[{Sun et~al.(2022)Sun, Shao, Wu, Shao and Fainman}]{sun2022reliable}
\bibinfo{author}{Sun, W.}, \bibinfo{author}{Shao, H.}, \bibinfo{author}{Wu,
  T.}, \bibinfo{author}{Shao, F.}, \bibinfo{author}{Fainman, E.Z.},
  \bibinfo{year}{2022}.
\newblock \bibinfo{title}{Reliable location of automatic vehicle identification
  sensors to recognize origin-destination demands considering sensor failure}.
\newblock \bibinfo{journal}{Transportation research part C: emerging
  technologies} \bibinfo{volume}{136}, \bibinfo{pages}{103551}.
%Type = Article
\bibitem[{Tang et~al.(2021)Tang, Cao, Chen, Yao, Tan and Sun}]{tang2021dynamic}
\bibinfo{author}{Tang, K.}, \bibinfo{author}{Cao, Y.}, \bibinfo{author}{Chen,
  C.}, \bibinfo{author}{Yao, J.}, \bibinfo{author}{Tan, C.},
  \bibinfo{author}{Sun, J.}, \bibinfo{year}{2021}.
\newblock \bibinfo{title}{Dynamic origin-destination flow estimation using
  automatic vehicle identification data: A 3d convolutional neural network
  approach}.
\newblock \bibinfo{journal}{Computer-Aided Civil and Infrastructure
  Engineering} \bibinfo{volume}{36}, \bibinfo{pages}{30--46}.
%Type = Article
\bibitem[{Veli{\v{c}}kovi{\'c} et~al.(2017)Veli{\v{c}}kovi{\'c}, Cucurull,
  Casanova, Romero, Lio and Bengio}]{velivckovic2017graph}
\bibinfo{author}{Veli{\v{c}}kovi{\'c}, P.}, \bibinfo{author}{Cucurull, G.},
  \bibinfo{author}{Casanova, A.}, \bibinfo{author}{Romero, A.},
  \bibinfo{author}{Lio, P.}, \bibinfo{author}{Bengio, Y.},
  \bibinfo{year}{2017}.
\newblock \bibinfo{title}{Graph attention networks}.
\newblock \bibinfo{journal}{arXiv preprint arXiv:1710.10903} .
%Type = Article
\bibitem[{Wang et~al.(2019a)Wang, Zheng, Ye, Gan, Li, Song, Zhou, Ma, Yu, Gai
  et~al.}]{wang2019deep}
\bibinfo{author}{Wang, M.}, \bibinfo{author}{Zheng, D.}, \bibinfo{author}{Ye,
  Z.}, \bibinfo{author}{Gan, Q.}, \bibinfo{author}{Li, M.},
  \bibinfo{author}{Song, X.}, \bibinfo{author}{Zhou, J.}, \bibinfo{author}{Ma,
  C.}, \bibinfo{author}{Yu, L.}, \bibinfo{author}{Gai, Y.}, et~al.,
  \bibinfo{year}{2019}a.
\newblock \bibinfo{title}{Deep graph library: A graph-centric,
  highly-performant package for graph neural networks}.
\newblock \bibinfo{journal}{arXiv preprint arXiv:1909.01315} .
%Type = Inproceedings
\bibitem[{Wang et~al.(2019b)Wang, Ji, Shi, Wang, Ye, Cui and
  Yu}]{wang2019heterogeneous}
\bibinfo{author}{Wang, X.}, \bibinfo{author}{Ji, H.}, \bibinfo{author}{Shi,
  C.}, \bibinfo{author}{Wang, B.}, \bibinfo{author}{Ye, Y.},
  \bibinfo{author}{Cui, P.}, \bibinfo{author}{Yu, P.S.}, \bibinfo{year}{2019}b.
\newblock \bibinfo{title}{Heterogeneous graph attention network}, in:
  \bibinfo{booktitle}{The world wide web conference}, pp.
  \bibinfo{pages}{2022--2032}.
%Type = Inproceedings
\bibitem[{Wang and Zhang(2022)}]{wang2022powerful}
\bibinfo{author}{Wang, X.}, \bibinfo{author}{Zhang, M.}, \bibinfo{year}{2022}.
\newblock \bibinfo{title}{How powerful are spectral graph neural networks}, in:
  \bibinfo{booktitle}{International Conference on Machine Learning},
  \bibinfo{organization}{PMLR}. pp. \bibinfo{pages}{23341--23362}.
%Type = Article
\bibitem[{Xiong et~al.(2020)Xiong, Ozbay, Jin and Feng}]{xiong2020dynamic}
\bibinfo{author}{Xiong, X.}, \bibinfo{author}{Ozbay, K.}, \bibinfo{author}{Jin,
  L.}, \bibinfo{author}{Feng, C.}, \bibinfo{year}{2020}.
\newblock \bibinfo{title}{Dynamic origin--destination matrix prediction with
  line graph neural networks and kalman filter}.
\newblock \bibinfo{journal}{Transportation Research Record}
  \bibinfo{volume}{2674}, \bibinfo{pages}{491--503}.
%Type = Article
\bibitem[{Zhang et~al.(2018)Zhang, Yuan, Zeng, Li and Wei}]{zhang2018missing}
\bibinfo{author}{Zhang, Q.}, \bibinfo{author}{Yuan, Q.}, \bibinfo{author}{Zeng,
  C.}, \bibinfo{author}{Li, X.}, \bibinfo{author}{Wei, Y.},
  \bibinfo{year}{2018}.
\newblock \bibinfo{title}{Missing data reconstruction in remote sensing image
  with a unified spatial--temporal--spectral deep convolutional neural
  network}.
\newblock \bibinfo{journal}{IEEE Transactions on Geoscience and Remote Sensing}
  \bibinfo{volume}{56}, \bibinfo{pages}{4274--4288}.
%Type = Article
\bibitem[{Zhang et~al.(2020)Zhang, Li, Lin and Wang}]{zhang2020network}
\bibinfo{author}{Zhang, Z.}, \bibinfo{author}{Li, M.}, \bibinfo{author}{Lin,
  X.}, \bibinfo{author}{Wang, Y.}, \bibinfo{year}{2020}.
\newblock \bibinfo{title}{Network-wide traffic flow estimation with
  insufficient volume detection and crowdsourcing data}.
\newblock \bibinfo{journal}{Transportation Research Part C: Emerging
  Technologies} \bibinfo{volume}{121}, \bibinfo{pages}{102870}.
%Type = Inproceedings
\bibitem[{Zhao et~al.(2021)Zhao, Wang, Shi, Hu, Song and
  Ye}]{zhao2021heterogeneous}
\bibinfo{author}{Zhao, J.}, \bibinfo{author}{Wang, X.}, \bibinfo{author}{Shi,
  C.}, \bibinfo{author}{Hu, B.}, \bibinfo{author}{Song, G.},
  \bibinfo{author}{Ye, Y.}, \bibinfo{year}{2021}.
\newblock \bibinfo{title}{Heterogeneous graph structure learning for graph
  neural networks}, in: \bibinfo{booktitle}{Proceedings of the AAAI conference
  on artificial intelligence}, pp. \bibinfo{pages}{4697--4705}.
%Type = Article
\bibitem[{Zhou and Mahmassani(2007)}]{zhou2007structural}
\bibinfo{author}{Zhou, X.}, \bibinfo{author}{Mahmassani, H.S.},
  \bibinfo{year}{2007}.
\newblock \bibinfo{title}{A structural state space model for real-time traffic
  origin--destination demand estimation and prediction in a day-to-day learning
  framework}.
\newblock \bibinfo{journal}{Transportation Research Part B: Methodological}
  \bibinfo{volume}{41}, \bibinfo{pages}{823--840}.
%Type = Article
\bibitem[{Zou and Chen(2020)}]{zou2020resilience}
\bibinfo{author}{Zou, Q.}, \bibinfo{author}{Chen, S.}, \bibinfo{year}{2020}.
\newblock \bibinfo{title}{Resilience modeling of interdependent
  traffic-electric power system subject to hurricanes}.
\newblock \bibinfo{journal}{Journal of Infrastructure Systems}
  \bibinfo{volume}{26}, \bibinfo{pages}{04019034}.

\end{thebibliography}

% \end{linenumbers}
\end{document}